\documentclass[runningheads]{llncs}

\usepackage{eccv}

\usepackage{eccvabbrv}

\usepackage{graphicx}
\usepackage{booktabs}

\newcommand*{\crosssymbol}{%
    \text{%
      \raise 1ex\hbox{%
        \rlap{\vrule height.2pt depth.2pt width .75ex}%
        \hbox to .75ex{\hss\vrule height .5ex depth 1ex\hss}%
      }%
    }%
}
\usepackage{graphbox}

\usepackage[accsupp]{axessibility}  

\usepackage{hyperref}

\usepackage{orcidlink}

\begin{document}


\title{Image Translation with Kernel Prediction Networks for Semantic Segmentation} 

\titlerunning{DA-KPN}

\author{Cristina Mata\inst{1} \thanks{Work performed as part of Microsoft Internship} \and
Michael Ryoo\inst{1} \and
Henrik Turbell\inst{2}}

\authorrunning{C.Mata et al.}

\institute{Stony Brook University, Stony Brook, NY 11794, USA\\
\email{\{cfmata, mryoo\}@cs.stonybrook.edu}\\
\and
Microsoft Corporation, Redmond, WA 98052, USA\\
\email{henrik.turbell@skype.net}}

\maketitle

\begin{abstract}
  Semantic segmentation relies on many dense pixel-wise annotations to achieve the best performance, but owing to the difficulty of obtaining accurate annotations for real world data, practitioners train on large-scale synthetic datasets. Unpaired image translation is one method used to address the ensuing domain gap by generating more realistic training data in low-data regimes. Current methods for unpaired image translation train generative adversarial networks (GANs) to perform the translation and enforce pixel-level semantic matching through cycle consistency. These methods do not guarantee that the semantic matching holds, posing a problem for semantic segmentation where performance is sensitive to noisy pixel labels. We propose a novel image translation method, Domain Adversarial Kernel Prediction Network (DA-KPN), that \textbf{guarantees} semantic matching between the synthetic label and translation. DA-KPN estimates pixel-wise input transformation parameters of a lightweight and simple translation function. To ensure the pixel-wise transformation is realistic, DA-KPN uses multi-scale discriminators to distinguish between translated and target samples. We show DA-KPN outperforms previous GAN-based methods on syn2real benchmarks for semantic segmentation with limited access to real image labels and achieves comparable performance on face parsing.
\end{abstract}

\section{Introduction}
\label{sec:intro}

\par Synthetic data is often used to train models for dense prediction tasks, leading to a drop in performance when those models are applied to real-world data. Unpaired image translation offers a way to circumvent this problem by first translating synthetic data into the real domain, then training models for downstream tasks using the translated images and source labels.
\par Current unpaired image translation methods use generative adversarial networks (GANs)\cite{Zhu17} to learn a mapping from samples in the source domain to the target domain. GANs are able to learn complex texture mappings between domains but are prone to changing the semantic structure of images \cite{Kim22}. This may be desirable for tasks such as image synthesis \cite{Kim22} but is detrimental to models requiring pixel-wise labels, since there is no guarantee that the synthetic label still corresponds to the translation at a pixel level. 
\begin{figure*}
  \centering
    \includegraphics[width=\textwidth]{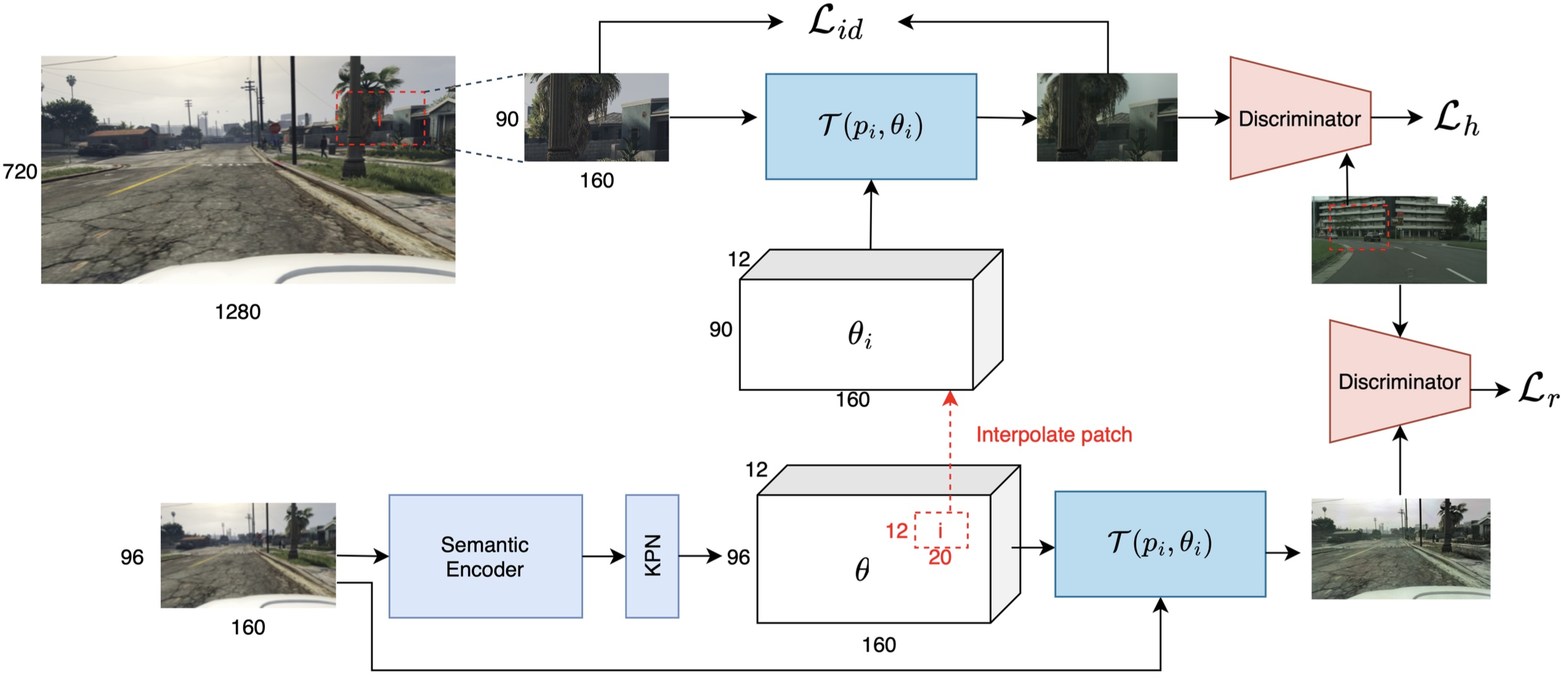}
  \caption{DA-KPN's overall architecture: A semantic encoder first extracts features from a low resolution source image which are used to predict pixel-wise transformation parameters with a Kernel Prediction Network (KPN). A 12 $\times$ 20 spatial patch of parameters is upsampled to 90 $\times$ 160 and used to transform the corresponding high resolution patch. Finally a multi-scale discriminator distinguishes between transformed and real samples at both the patch and global image level.}
  \label{fig:overall_arch}
\end{figure*}
\par Cycle consistency \cite{Zhu17} was introduced as a constraint on the mapping from source to target: it constrains the mapping back from target to source to be the identity, ensuring that translations do not vastly diverge from their inputs. Regularization losses can also constrain the translation \cite{Ko22}, but neither of the above ensures that the geometry of the objects in the source image remain unchanged. 
\par Disentangling the latent space of GANs into style and content features is another approach that has seen success in image translation \cite{Gabbay21}. These methods often require two separate encoders \cite{Gabbay21}, leading to increased memory usage which may be prohibitive in certain settings. Noisy labels are still a byproduct of these translations and have not been adequately addressed by previous methods.
\par Our goal is to learn a lightweight photorealistic image translation function whose translations do not result in label noise on downstream segmentation tasks. We propose an approach called Domain Adversarial Kernel Prediction Network (DA-KPN) that performs translations efficiently using simple pixel-wise convolutions whose kernel parameters vary depending on the semantic content of each pixel. There are several advantages to DA-KPN which are particularly favorable for downstream segmentation:
\begin{enumerate}
    \item Geometry of input image is exactly preserved, preventing semantic flipping.
    \item Pixel-wise transformation is fast and scales to high resolution images.
    \item Translation is interpretable and depends on pixel-wise semantic identity in addition to underlying data distribution.
\end{enumerate}
\par To ensure the translations depend on the semantic features that best benefit the downstream task, DA-KPN uses a semantic segmentation network to extract pixel-wise features with a low spatial resolution (Figure \ref{fig:overall_arch}). Each feature is mapped to a set of parameters with a Kernel Prediction Network. Feature encoding and kernel prediction is performed on low resolution inputs to keep memory usage low. The parameters are used as part of a lightweight translation function to transform the corresponding pixel in the input. To allow transformation of high resolution images, the parameters are upsampled to the full resolution prior to transformation. The transformation is simple, fast and parallelizable: it is composed of a linear color transformation, symmetric Gaussian blur, and scalable noise addition. DA-KPN uses a multi-scale discriminator that distinguishes between patches and whole images from translated and real domains in order to learn realistic transformation parameters.
\par We evaluate DA-KPN's performance on translating synthetic images into the real domain on two syn2real benchmarks: semantic segmentation using GTA$\rightarrow$ Cityscapes \cite{Richter16} \cite{Cordts16} and face parsing using Face Synthetics$\rightarrow$CelebAMask-HQ \cite{Wood21} \cite{Lee20a}. The final downstream segmentation network has access only to source annotations. We find that a network trained on the GTA translations generated by DA-KPN outperform networks trained on translations from GAN-based methods CycleGAN \cite{Zhu17} and VSAIT \cite{Theiss22}. Similarly, a network trained on DA-KPN translations of FaceSynthetics outperforms the same network trained on CycleGAN translations and performs comparably to VSAIT. DA-KPN achieves these results while requiring much less time and memory to train. Our results indicate that DA-KPN offers a strong alternative image translation method to GANs that is well suited to downstream segmentation tasks, especially in low-compute resource settings.

\section{Related Work}

\subsection{Kernel Prediction Networks}

\par Kernel prediction has been used in blind image super-resolution methods where blurry images are modeled as a high resolution image convolved with an unknown blur kernel \cite{Fang22} \cite{Zheng22} \cite{Zhang19} \cite{Yue22} \cite{Emad22} \cite{Zhou19} \cite{Gu19}. The goal for these approaches is to explicitly solve for possibly non-Gaussian blur kernels using the low resolution input, whereas in our work we parameterize our blur kernel as a Gaussian with a single parameter. Closer to our network, \cite{Liang21a} uses features from a low resolution image to generate spatially varying blur kernel parameters. \cite{Liang21b} introduces a flow-based prior to guide the kernel estimation process. \cite{Kim21} takes features from low-resolution inputs to predict blur kernels and then adaptively changes the kernels for the high resolution input.
\par In high resolution image-to-image translation, kernel parameters have been estimated specifically for use in the instance normalization layer of generative adversarial network models \cite{Ho22}. AdaCM \cite{Lin23} predicts the parameters of an MLP that approximates a 3D Look Up Table for color stylization. Our work similarly models color using an affine transformation but in addition models blur and noise.

\subsection{Unpaired Image Translation}
In unpaired image translation, a model takes an image in the source domain as input and outputs a transformed version of the image in the target domain. Most methods for unpaired image translation focus on learning a mapping from source to target directly from the data. They differ in the ways that they impose constraints on the translation. CycleGAN \cite{Zhu17} is a method still used today in which a cycle consistency loss enforces the inverse mapping from translation to input to approximate the original input. Recently a class of methods have been introduced which focus on disentangling content/structure and texture in input representations in order to handle them differently in the translation or enable user-guided stylistic changes \cite{Chen22}. CUT \cite{Park20} is an example of this where corresponding patches in the input and translation are forced to share mutual information through a contrastive loss. While GAN-based methods excel at learning each domain's feature distributions, the translation often exhibits semantic flipping: pixels in the source image may be transformed to different objects, rendering semantic labels for segmentation incorrect. Recently VSAIT \cite{Theiss22} was introduced to deal with semantic flipping in the translation by learning a hash map of features into hypervector space and then enforcing a cycle consistency loss in that space. In contrast to previous methods, DA-KPN learns a small set of parameters for a simple translation function applied directly to the input, which prevents any changes in structure. Multi-Curve Translator (MCT) \cite{Song22} also predicts the parameters for a pixel-wise transformation function but uses a curve-based look up table as a translation function, whereas we use a simpler transformation based on convolutions.
\par Previous unpaired image translation methods have focused on generating images that look ``natural" but which may not be optimal for downstream segmentation tasks due to a lack of realistic noise distribution. Since it is common practice to translate synthetic to real images for semantic segmentation training, the utility of translation methods is best evaluated by training on translations and evaluating semantic segmentation metrics on target data. However there is a lack of reporting on translation methods' quantitative performance for segmentation. SRUNIT \cite{Jia21} trained a Deeplab \cite{Chen17} segmentation network on GTA translations and labeled Cityscapes target images and reports segmentation metrics on only 500 source images set aside for evaluation. VSAIT \cite{Theiss22} reports Deeplab-Scores on the GTA$\rightarrow$Cityscapes task. CycleGAN \cite{Zhu17}, GcGAN \cite{Fu19} and CUT \cite{Park20} evaluate FCN/DRN-Scores on the Cityscapes label$\rightarrow$photo task, but this task is very different from our focus since it does not have unmatched semantics. DRIT \cite{Lee20b} and MUNIT \cite{Huang18} only report qualitative syn2real results. We aim to fill the gap on segmentation in the literature.

\section{Method}

\subsubsection{Overview} DA-KPN's overall architecture, shown in Figure \ref{fig:overall_arch}, consists of a semantic feature encoder, kernel prediction module and multi-scale discriminator. Source images are resized to a low resolution and fed to the encoder to extract semantic information while retaining the spatial resolution of the image. Extracting features at low resolution prevents high memory usage and preserves pixel-level semantics. The features are passed to the kernel prediction module that outputs a set of parameters for every pixel in the feature. These parameters are upsampled to the high resolution of the source image. The translation is applied to the full resolution source image using the upsampled parameters. During training, a discriminator distinguishes between translated patches and random target patches to encourage local photorealistic translation. To ensure that translations are globally photorealistic, the full translated image and random target images are input to the discriminator.

\subsection{Translation Function}
\par
\begin{figure}
  \centering
    \includegraphics[width=0.8\textwidth]{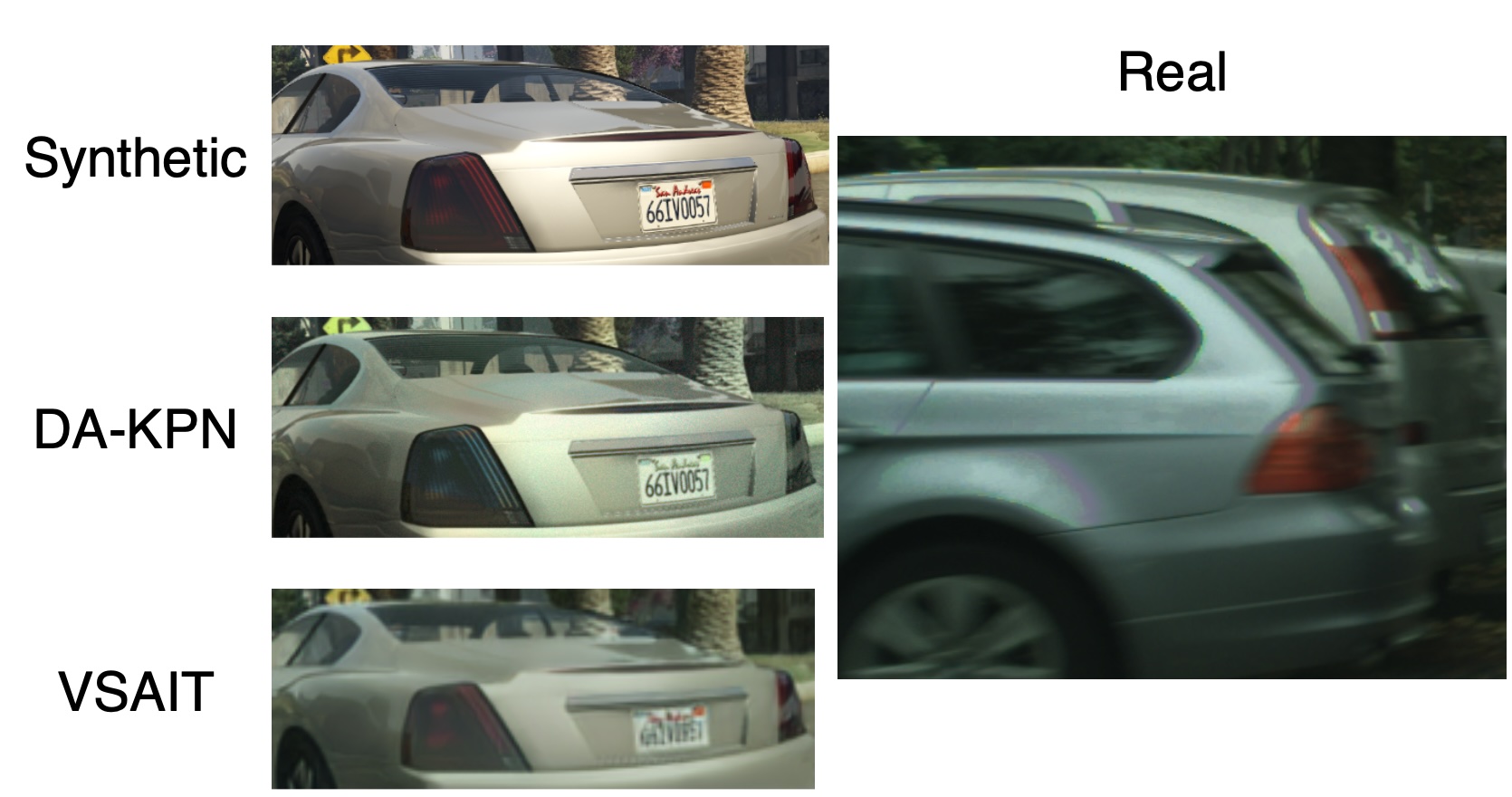}
  \caption{Crop from a real image versus synthetic image and translations. Real images have color artifacts (such as purple and green in the shadow in the car) and random noise, which DA-KPN explicitly models.}
  \label{fig:crops}
\end{figure}

 Based on local inspection of synthetic and real images, shown in Figure \ref{fig:crops}, the syn2real domain gap can be characterized by color artifacts, random noise and motion blur from the object or camera. DA-KPN seeks to explicitly model such domain differences in a lightweight translation function $\mathcal{T}$. Specifically, $\mathcal{T}$ is defined as a composition of a linear color transformation, $25 \times 25$ Gaussian kernel convolution and additive noise. 
\par Formally, $\forall p_{i} \in \mathcal{I}$,
\begin{equation}
    \mathcal{T}(p_{i}; \theta_{i}) = n_{i} \times \mathcal{N}_{i} + k_{\sigma_{i}} * (w_{i} \times p_{i} + b_{i})
\label{eqn:transformation}
\end{equation}
where $p_{i}$ is a pixel at location $i \in (H,W,3)$, $\mathcal{I} \in \mathbb{R}^{H \times W \times 3}$ is the image and the parameters $\theta_{i} = [w_{i}; b_{i}; \sigma_{i}; n_{i}]$ are defined per-pixel. $w_{i} \times p_{i}$ refers to element-wise multiplication between weight $w_{i}$ and pixel $p_{i}$: each channel for each pixel is multiplied with separate weight. $w_{i}$, $b_{i}$ and $p_{i}$ are all vectors of length 3 corresponding to the number of channels in the input image. The blur kernel $k_{{\sigma}_{i}}$ is a 2D Gaussian with kernel size $25 \times 25$ calculated for pixel $i$ as 
\begin{equation}
k_{\sigma_{i}} = \frac{1}{2\pi\sigma_{i}^{2}}e^{-\frac{(x^{2}+y^{2})}{2\sigma_{i}^{2}}} 
\end{equation}
with $(x,y)$ denoting a position in the blur matrix. The large kernel size allows flexible application of small to large blurs. DA-KPN uses a symmetric Gaussian kernel to model blur to prevent structural changes to the image.
\par The constant noise variable $\mathcal{N}$ is generated with the same dimensions as $\mathcal{I}$ once prior to training. Each pixel's noise is randomly sampled from a Gaussian centered at 0 with variance 1, and the noise is clamped to be in the range [0, 1]. 
\par DA-KPN's translation function targets artifacts of real images often absent from synthetics: motion blur, poor lighting and random noise. Neighboring pixels in the image are able to undergo different transformations depending on their semantic identity. Although this work focuses on the syn2real domain gap, an added benefit of DA-KPN's overall architecture is its flexibility to allows different translation functions to be swapped out in case of different source and target domains. 

\subsection{Losses}

\par To ensure the translation is photorealistic at a local and global level, DA-KPN uses a multiscale discriminator that distinguishes between (a) patches of high resolution real and translated samples and (b) low resolution real and translated images. Formally, the high resolution loss propagated from discriminator $\mathcal{D}$ is
\begin{equation}
    \mathcal{L}_{h} = -\log(1-\mathcal{D}(P_{t})) - \log(\mathcal{D}(P_{r}))
\end{equation}
where $P_{t}$ and $P_{r}$ are patches from high resolution translated and real samples. Similarly the low resolution loss operates over the low resolution translated and real images $\mathcal{I}_{t}^{l}$ and $\mathcal{I}_{r}^{l}$:
\begin{equation}
    \mathcal{L}_{r} = -\log(1-\mathcal{D}({\mathcal{I}_{t}^{l}}) - \log(\mathcal{D}(\mathcal{I}_{r}^{l}))
\end{equation}

To prevent extreme deviations in the translation from the source image DA-KPN employs a Mean Squared Error identity loss $\mathcal{L}_{id}$ in RGB space between the high resolution translated patch $P_{t}$ and original high resolution synthetic patch $P_{s}$. The final loss function is the sum of all of these:
\begin{equation}
    \mathcal{L} = \mathcal{L}_{h} + \mathcal{L}_{r} + \mathcal{L}_{id}
\end{equation}

\subsection{Implementation}

\par The semantic encoder is implemented as an FCN-Resnet50 \cite{Long15}. Prior to training DA-KPN, the encoder is pretrained for semantic segmentation on low resolution $96 \times 160$ labeled source images. During DA-KPN training the encoder is frozen and the rest of the model is finetuned. Finetuning the encoder at the same time as the discriminator without any additional constraints results in mode collapse in the encoder. 
\par DA-KPN's kernel prediction module is implemented as a 2D convolution with $n$ input channels and 12 output channels, where $n$ is the number of semantic classes in the dataset. The first 6 output channels are assigned to the weight and bias parameters for color translation, predicted for each input channel. Prior to applying the linear color translation, the input is converted to HSV color space, and converted back to RGB space after the translation. The next 3 channels correspond to the blur kernel $\sigma$ parameters for input RGB channels. The remaining channels correspond to the additive noise scaling for input RGB channels. The KPN is initialized to output parameters giving the identity translation. Each spatial feature in the KPN's output consists of the 12 parameters needed in Equation \ref{eqn:transformation} for that pixel. 
\par After using those parameters to translate the input, the translated image is passed to the multi-scale discriminator. The discriminator is implemented as a PatchGAN \cite{Isola17} with 3 layers and 64 hidden features. We use Adam optimizer with learning rate set to 1e-4 and batch size 8, alternating updates to the KPN and discriminator. We set the number of training iterations to 100,000. The code may be found at \url{https://github.com/cfmata/dakpn}.
\par The high resolution input images have spatial dimensions $720 \times 1280$ and the low resolution images are $96 \times 160$. During translation each dimension is divided by a factor of 8 to produce parameter patches of size $12 \times 20$ from the low resolution feature. These are upsampled to $90 \times 160$ and used to transform the corresponding $90 \times 160$ patch in the original synthetic input. Patches of size $90 \times 160$ are used from the real images for the discriminator.

\section{Experiments}

\par Our focus in this work is on improving the performance of dense prediction networks in settings with access to labeled synthetic data labels and limited access to labeled real data. We evaluate DA-KPN's translations based on the performance of semantic segmentation and face parsing models trained on the \emph{translated} source dataset. 

\begin{table}[t]
  \centering
  \caption{FCN-Scores reported on the GTA translations using an FCN-Resnet50 trained on real Cityscapes training set. We compare our DA-KPN to the state of the art based on their reported results and outperform previous methods. $\crosssymbol$ indicates the method is retrained.}
  \begin{tabular}{cccc}
    \toprule
    Translation Method & pxAcc & clsAcc &  \textbf{mIOU} \\
    \midrule
    DRIT \cite{Lee18} & 64.28 & 32.17 & 20.99 \\
    CUT \cite{Park20} & 64.59 & 32.19 & 20.35 \\
    CycleGAN$^{\crosssymbol}$ \cite{Zhu17} & \textbf{78.49} & 38.04 & 22.41 \\
    GcGAN \cite{Fu19} & 65.62 & 32.38 & 22.64 \\
    SRUNIT \cite{Jia21} & 67.21 & 32.97 & 22.69 \\
    VSAIT$^{\crosssymbol}$ \cite{Theiss22} & 74.36 & \underline{49.23} & \underline{32.42} \\
    DA-KPN-S (ours) & \underline{76.75} & \textbf{49.39} & \textbf{34.11} \\
    \bottomrule
  \end{tabular}
  \label{tab:fcn_scores_sota}
\end{table}

\begin{table}
  \centering
  \caption{Computation information on GTA$\rightarrow$Cityscapes: \# parameters (millions), single epoch train time and single image inference time.}
\begin{tabular}{|c|c|c|c|}
\hline
     \textbf{Method} & \textbf{Param (M)} & \textbf{Train (sec)} & \textbf{Inf. (sec)} \\
    CycleGAN & 28.3 & 11327 & 0.08 \\
    VSAIT & 65.5 & 20400 & 1.30 \\
    SRUNIT & 14.70 & 1248 & 0.235 \\
    CUT & 14.70 & 5992 & 0.24 \\
    DA-KPN (ours) & 2.8 & 4420 & 0.52 \\
  \hline
  \end{tabular}
  \label{tab:computation}
\end{table}

\begin{table*}[t]
  \centering
  \caption{Semantic segmentation results on Cityscapes by a an FCN-Resnet50 trained on GTA translations. DA-KPN-S has a source-pretrained encoder, while DA-KPN-T is target-pretrained. DA-KPN's translations improve the network's performance over GAN-based methods.}
  \begin{tabular}{c|ccc|ccc}
    \toprule
    Resolution & \multicolumn{3}{c|}{512 $\times$ 512} & \multicolumn{3}{c}{720 $\times$ 1280} \\
    \midrule
    Translation Method & pxAcc & clsAcc & \textbf{mIOU} & pxAcc & clsAcc & \textbf{mIOU} \\
    \midrule
    CycleGAN & 81.25 & 48.02 & 35.31 & 78.73 & 43.18 & 33.64 \\
    VSAIT & \textbf{82.71} & 48.79 & \underline{36.82} & 74.87 & 46.70 & \underline{35.23}\\
    DA-KPN-S (ours) & \underline{81.70} & \underline{48.98} & 36.41 & \underline{78.07} & \underline{48.46} & 35.17\\
    DA-KPN-T (ours) & 80.25 & \textbf{50.87} & \textbf{37.67} & \textbf{78.62} & \textbf{49.67} & \textbf{37.45} \\
    \midrule
    {\color{gray} Target} & {\color{gray}92.83} & {\color{gray}69.85} & {\color{gray}60.33} & {\color{gray}92.83} & {\color{gray}69.12} & {\color{gray}60.16} \\
    \bottomrule
  \end{tabular}
  \label{tab:segmentation_results_gta}
\end{table*}

\subsection{Semantic Segmentation}

\par We test whether DA-KPN is able to translate synthetic images from GTA \cite{Richter16} into realistic images emulating Cityscapes \cite{Cordts16} for the downstream task of semantic segmentation. GTA consists of 24,966 synthetic images of street scenes collected from a video game and comes with pixel-level semantic labels for 19 classes. Cityscapes is a real street-scene dataset with fine pixel-level annotations across 19 classes for 5,000 images. GTA$\rightarrow$Cityscapes is commonly used for syn2real image translation due to the overlapping class sets.
\par We train DA-KPN using all synthetic images from GTA and unlabeled real images from the Cityscapes training split. By default we pretrain the encoder on labeled source images and denote this model as DA-KPN-S. We also experiment with pretraining on the small amount of labeled target data available in the training set and denote this model as DA-KPN-T. We use a single NVIDIA RTX A5000 GPU with 24 GB of memory for training DA-KPN. After training we use DA-KPN to translate the entire GTA dataset. 
\par We first report FCN-Scores as described in \cite{Zhu17}: we train an FCN-Resnet50 on real images in the Cityscapes training set and evaluate this model on the GTA translations generated by DA-KPN. The FCN-Scores consist of the pixel accuracy, class accuracy and mean IOU (mIOU) over all semantic classes of the target dataset. Pixel accuracy is defined as the percentage of pixels in the entire image that are assigned the correct class. Class accuracy is the mean pixel accuracy across all classes. mIOU refers to the Jaccard index averaged over all classes. Our main focus is the mIOU metric because it captures the model's performance on small, uncommon objects that are most challenging to segment. We compare FCN-Scores for DA-KPN to the state of the art image translation methods in Table \ref{tab:fcn_scores_sota} and show that DA-KPN outperforms all other methods. Note that we retrained the baselines CycleGAN and VSAIT on unlabeled images from the full GTA dataset and use the training split of Cityscapes as required by those methods.
\par To test whether DA-KPN's translations are beneficial for semantic segmentation, we train an FCN-Resnet50 on the entire translated GTA dataset and test it on the validation split of Cityscapes. In Table \ref{tab:segmentation_results_gta} we evaluate on all 19 classes in the Cityscapes validation set consisting of 500 labeled images, resizing the images to either 512$\times$512 resolution or the full 720$\times$1280 resolution. DA-KPN-T outperforms both VSAIT and CycleGAN while DA-KPN-S is on par with the state of the art. We compare the computational efficiency of DA-KPN to baselines in Table \ref{tab:computation}, showing that DA-KPN requires less memory and is faster to train than CycleGAN and VSAIT. It is also able to perform inference faster than the current state of the art. 
\par Based on the quantitative results in Tables \ref{tab:fcn_scores_sota}–\ref{tab:segmentation_results_gta}, DA-KPN generates translations that are better suited for semantic segmentation rather than those generated by CycleGAN or VSAIT, particularly in cases when training on high resolution images is prohibitively expensive: DA-KPN-S/T gets good performance at 720 $\times$ 1280 resolution in Table \ref{tab:segmentation_results_gta}. When a small amount of labeled target data is available, DA-KPN can take advantage of it to gain a performance boost \ref{tab:segmentation_results_gta}.
\par
\begin{figure*}[t]
\tabcolsep 0.05cm
\noindent\makebox[\textwidth]{
\begin{tabular}{cccc}
  Source &
  \includegraphics[align=c,trim = 0mm 0mm 0mm 0mm, clip, width=3cm]{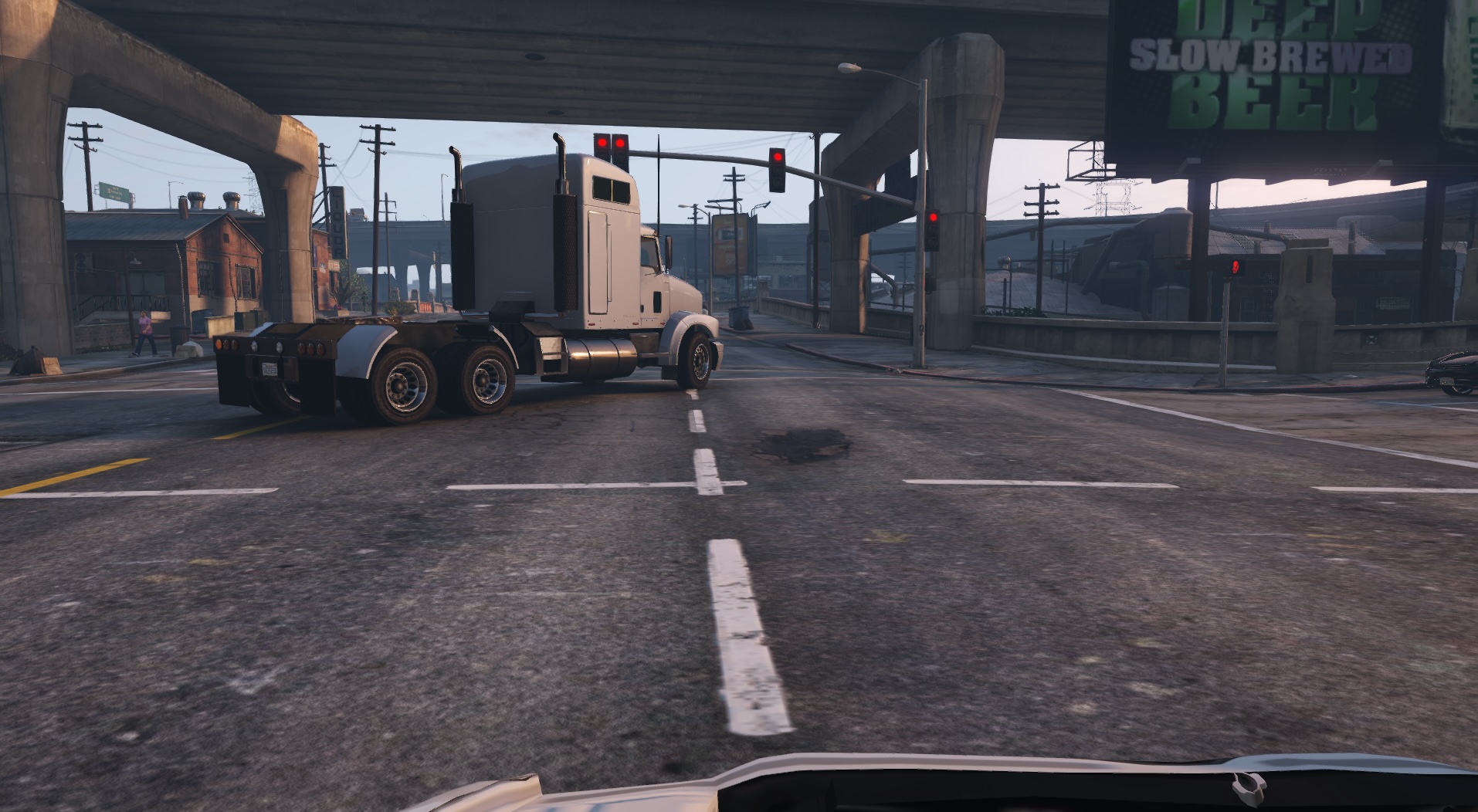} &
  \includegraphics[align=c,trim = 0mm 0mm 0mm 0mm, clip, width=3cm]{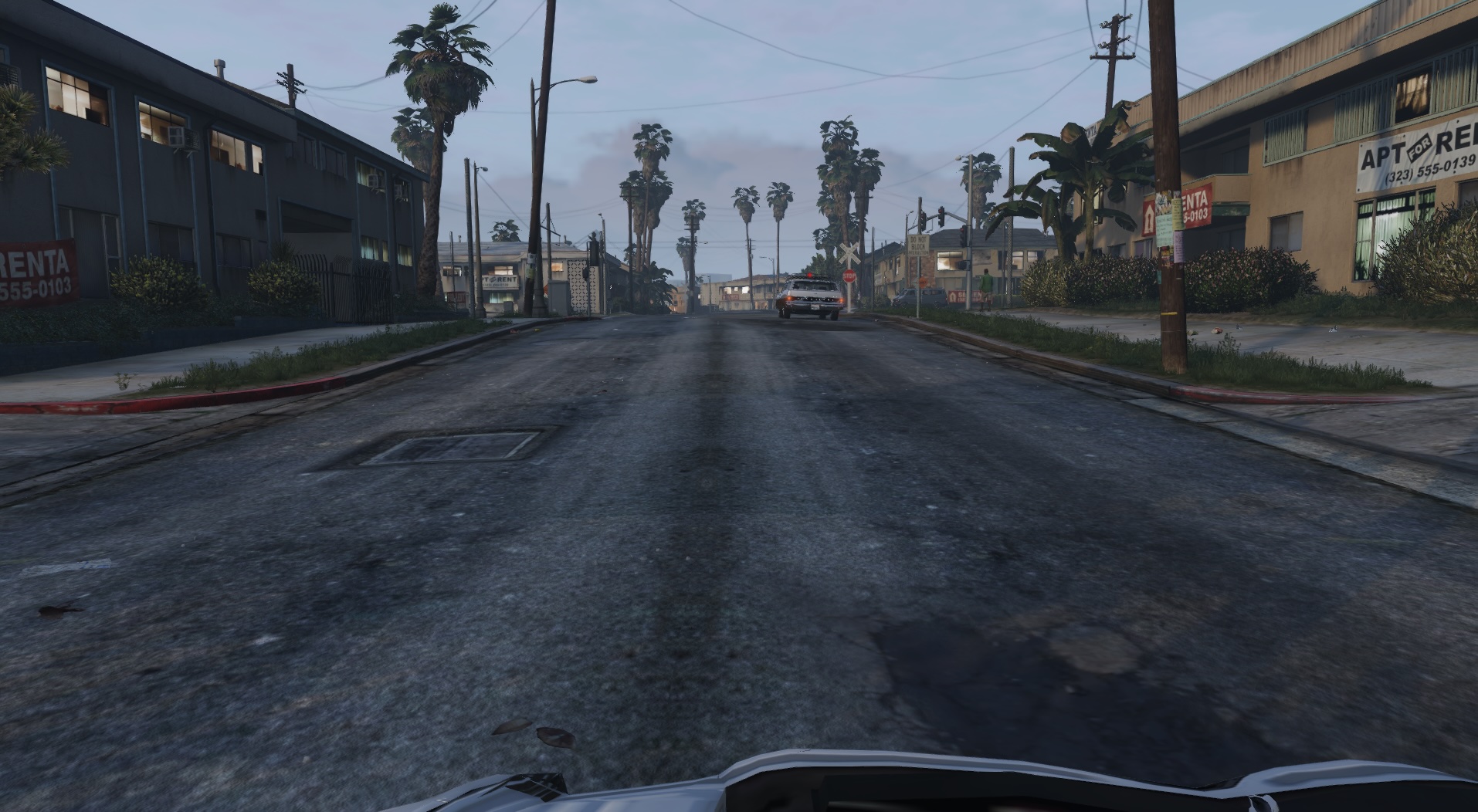} &
  \includegraphics[align=c,trim = 0mm 0mm 0mm 0mm, clip, width=3cm]{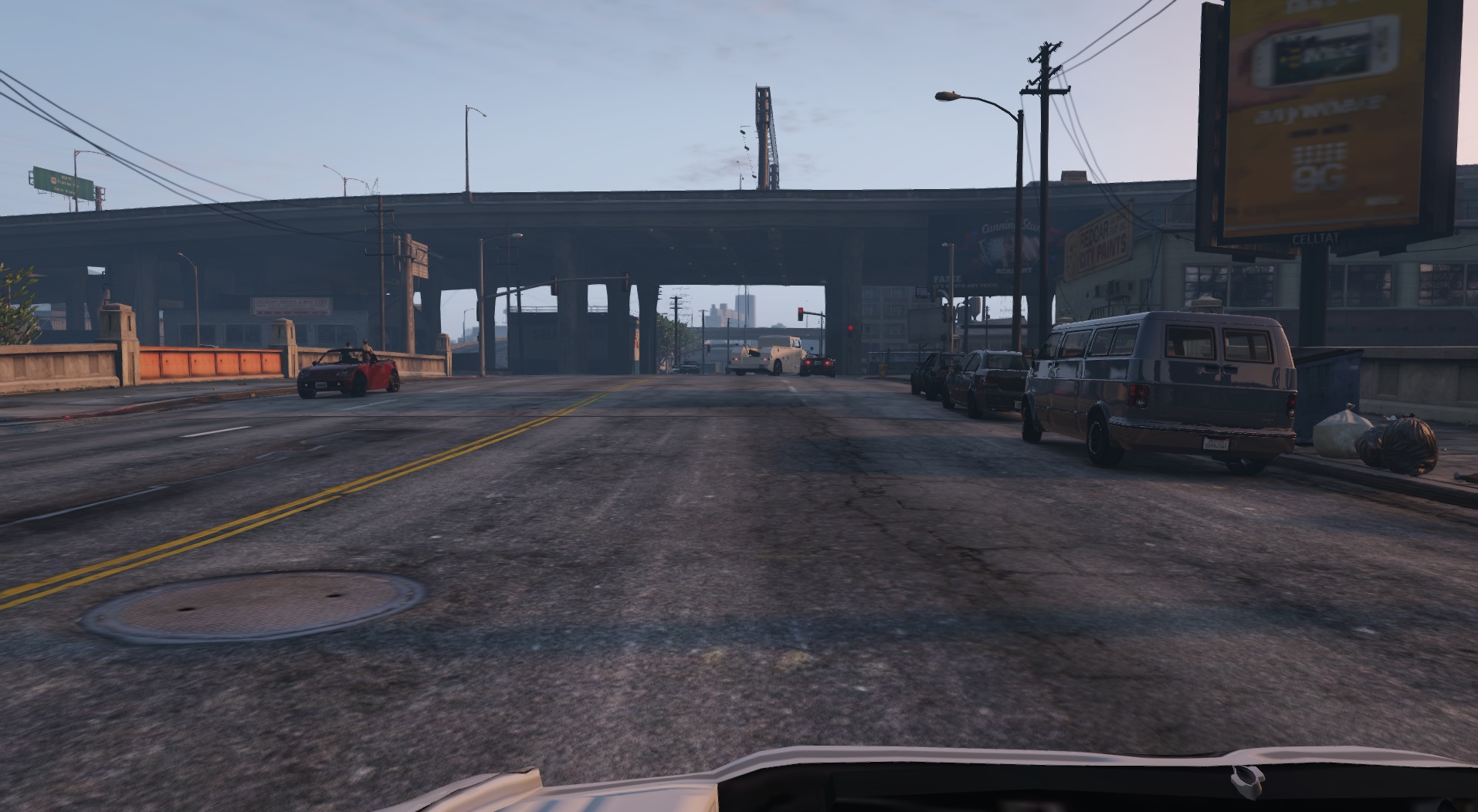} \\
  
  CycleGAN &
  \includegraphics[align=c,trim = 0mm 0mm 0mm 0mm, clip, width=3cm]{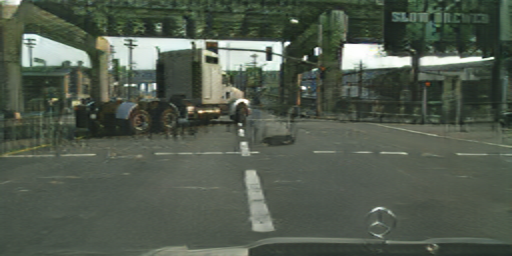} &
  \includegraphics[align=c,trim = 0mm 0mm 0mm 0mm, clip, width=3cm]{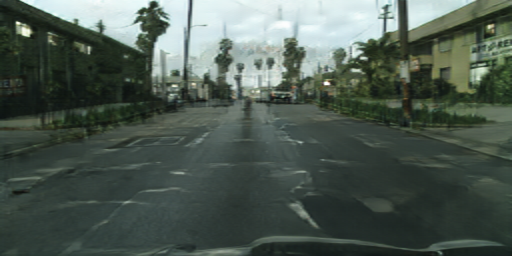} &
  \includegraphics[align=c,trim = 0mm 0mm 0mm 0mm, clip, width=3cm]{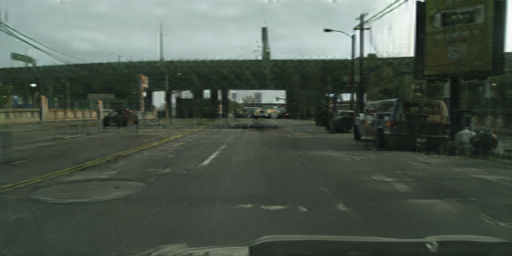} \\
  
  VSAIT &
  \includegraphics[align=c,trim = 0mm 0mm 0mm 0mm, clip, width=3cm]{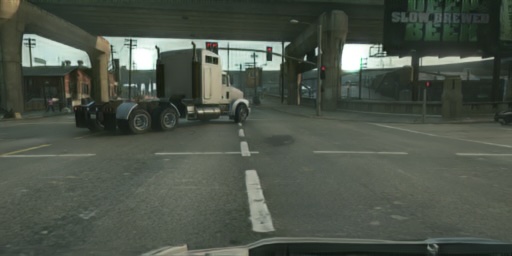} &
  \includegraphics[align=c,trim = 0mm 0mm 0mm 0mm, clip, width=3cm]{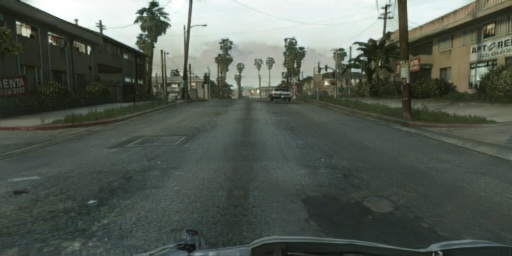} &
  \includegraphics[align=c,trim = 0mm 0mm 0mm 0mm, clip, width=3cm]{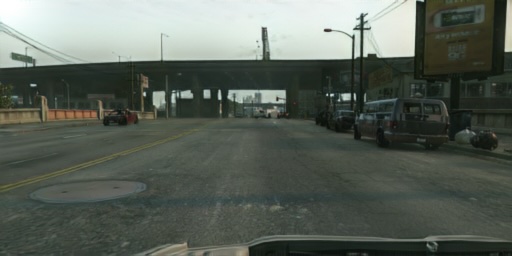} \\
  
  \textbf{DA-KPN} &
  \includegraphics[align=c,trim = 0mm 0mm 0mm 0mm, clip, width=3cm]{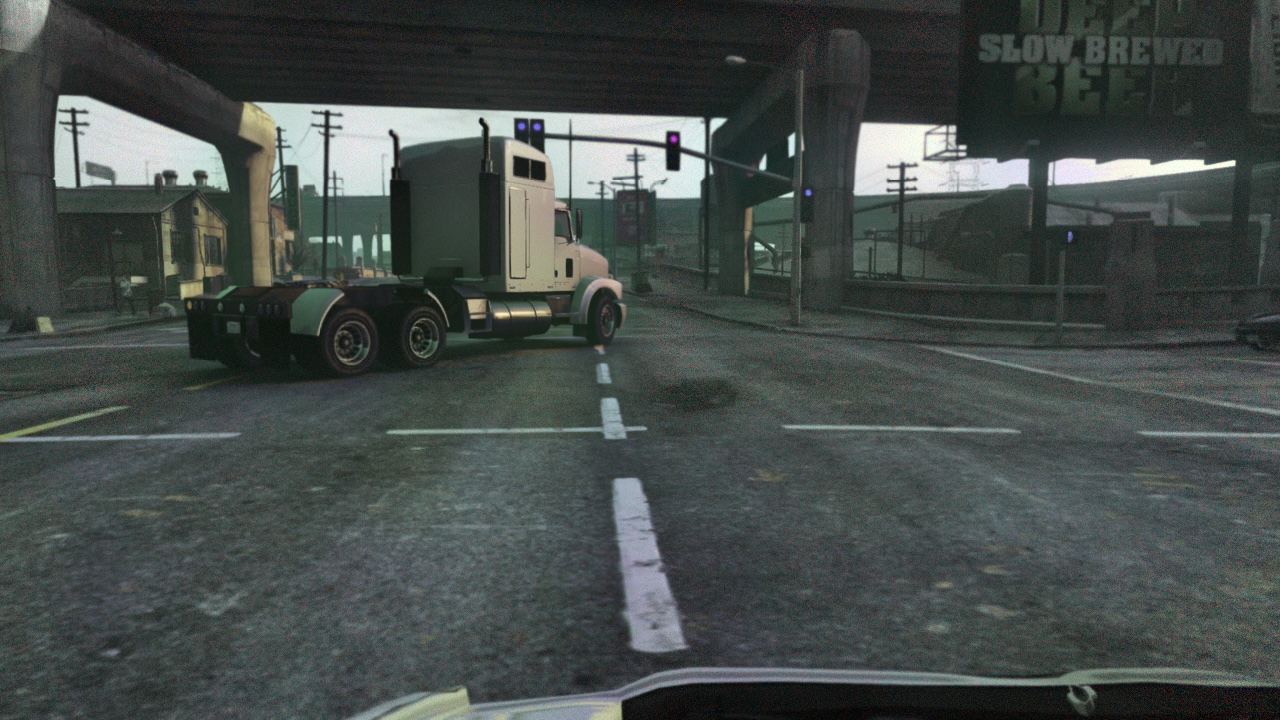} &
  \includegraphics[align=c,trim = 0mm 0mm 0mm 0mm, clip, width=3cm]{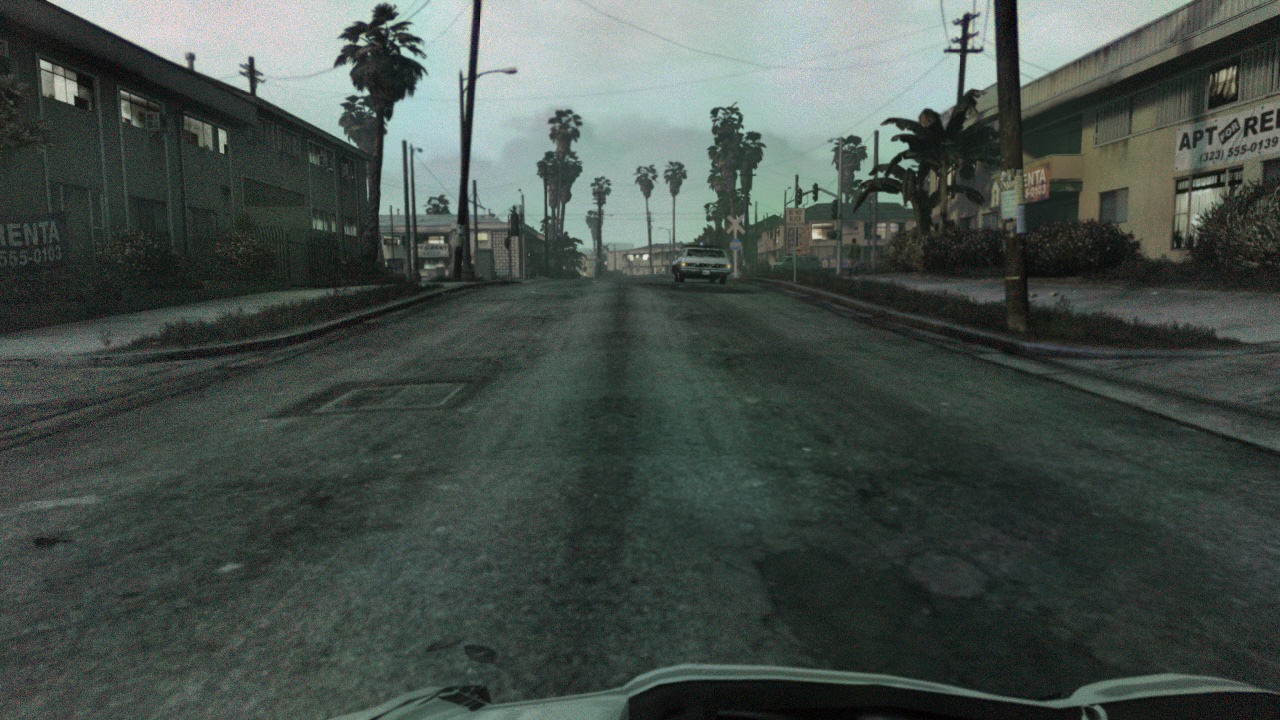} &
  \includegraphics[align=c,trim = 0mm 0mm 0mm 0mm, clip, width=3cm]{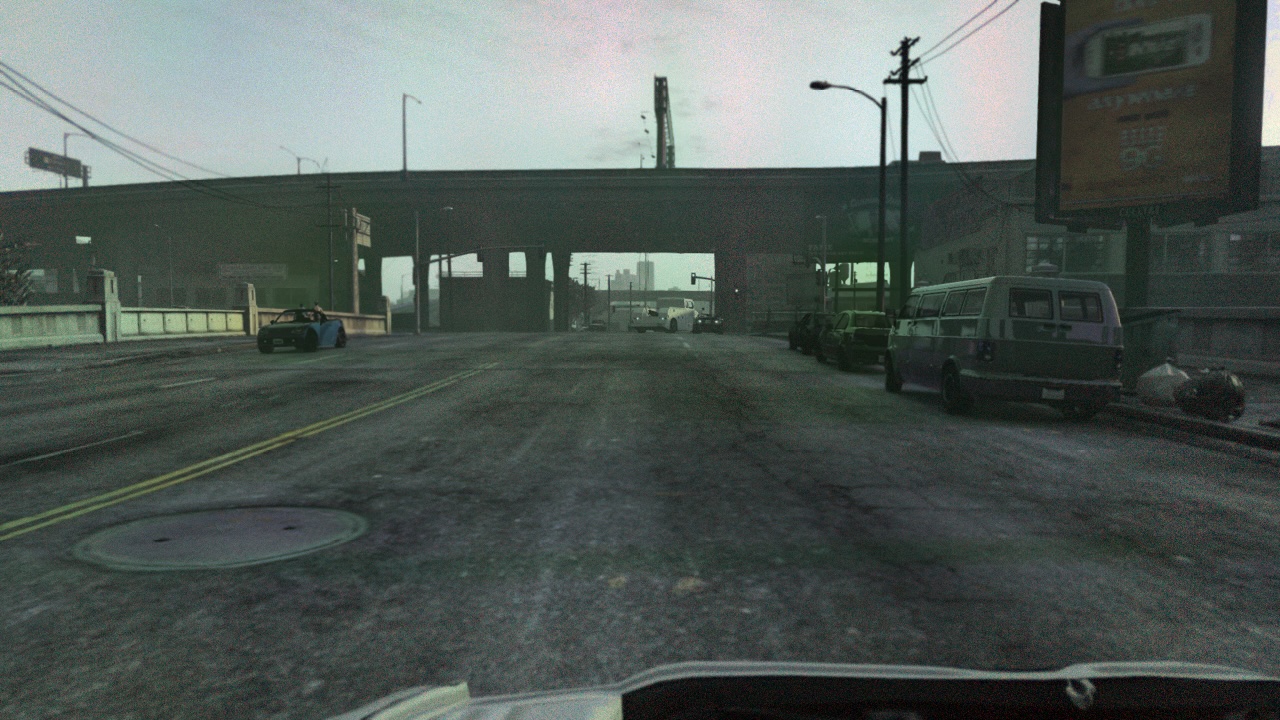} \\
  
  Target &
  \includegraphics[align=c,trim = 0mm 0mm 0mm 0mm, clip, width=3cm, height=1.8cm]{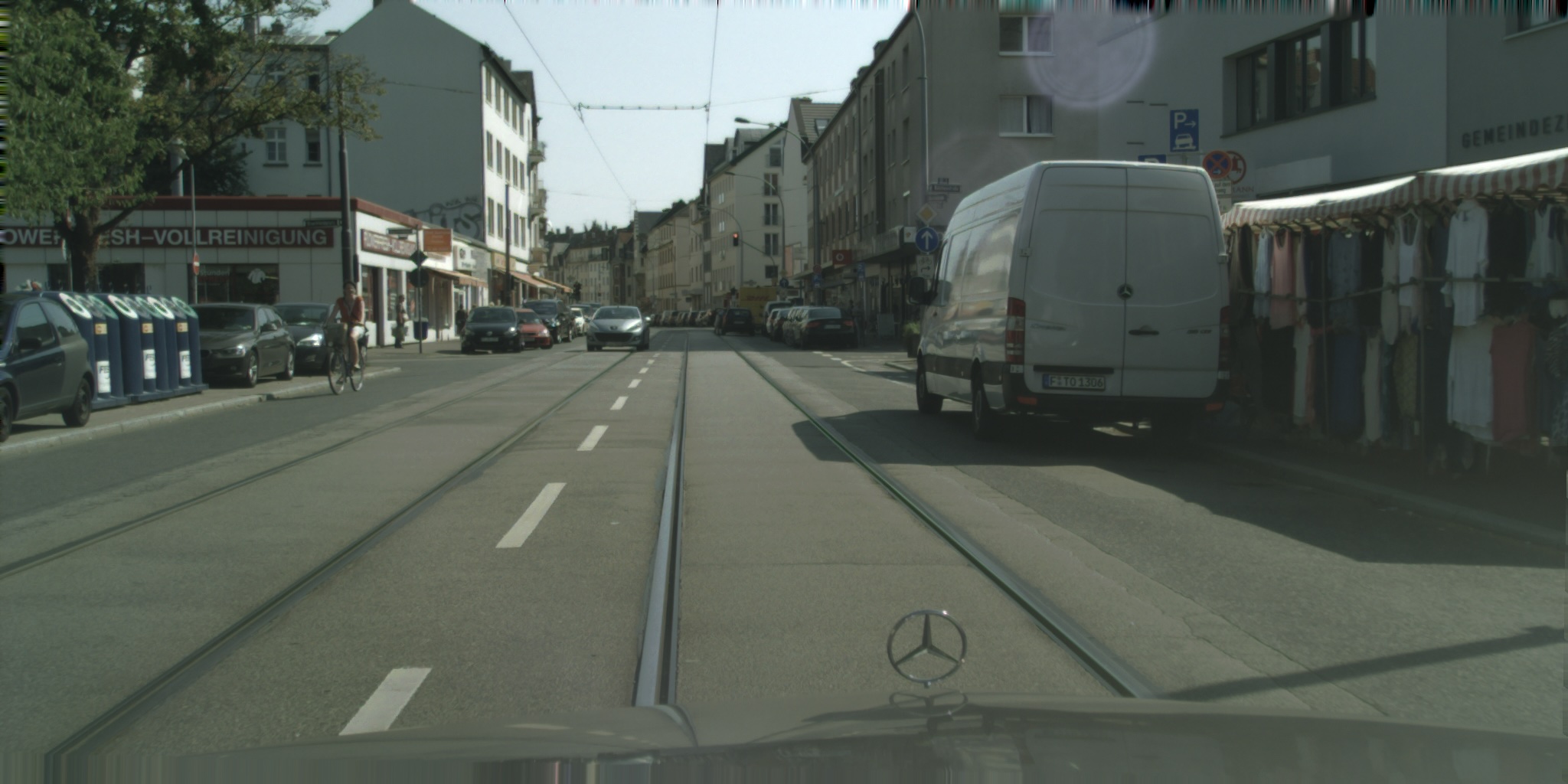} &
  \includegraphics[align=c,trim = 0mm 0mm 0mm 0mm, clip, width=3cm, height=1.8cm]{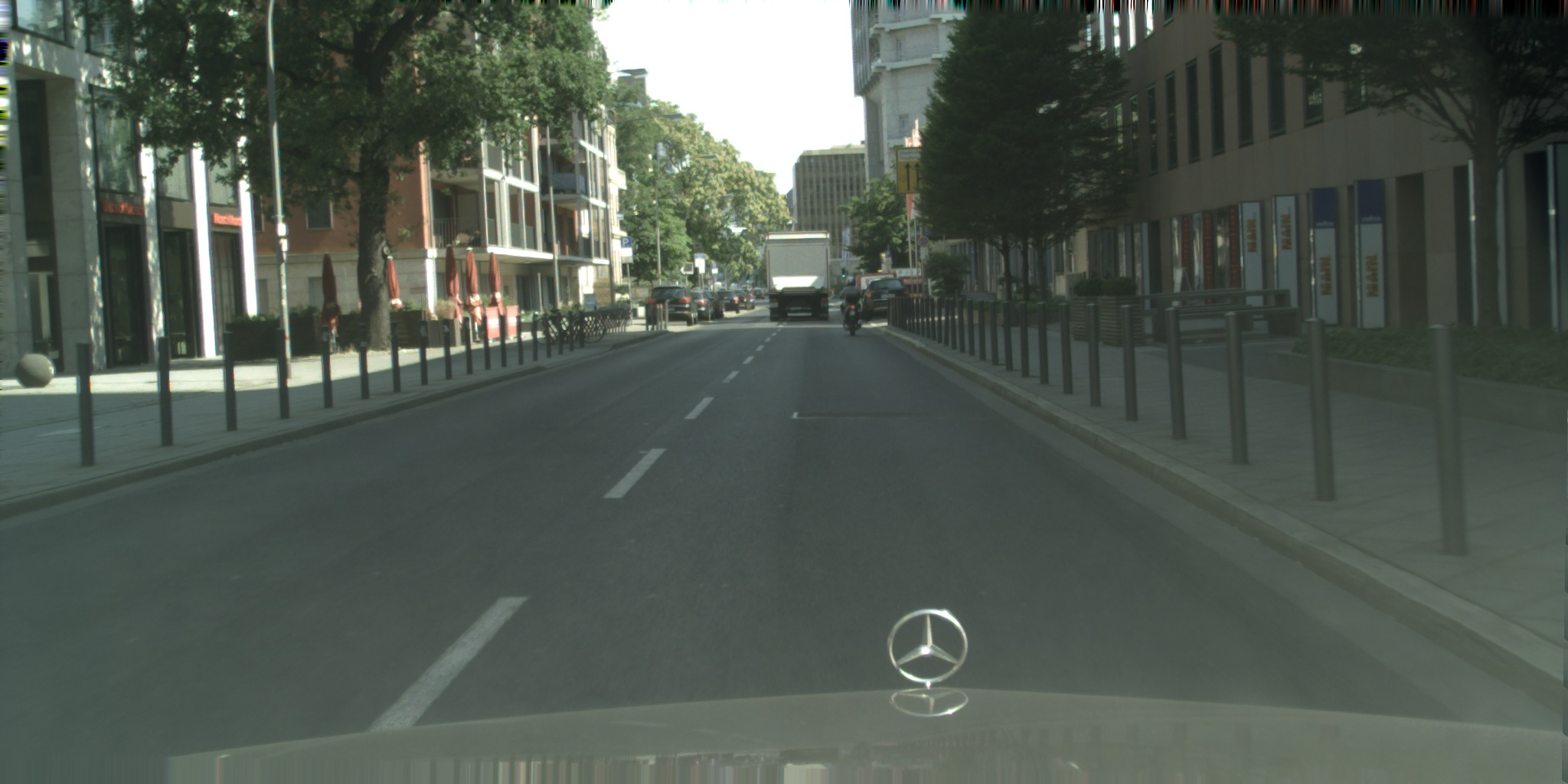} &
  \includegraphics[align=c,trim = 0mm 0mm 0mm 0mm, clip, width=3cm, height=1.8cm]{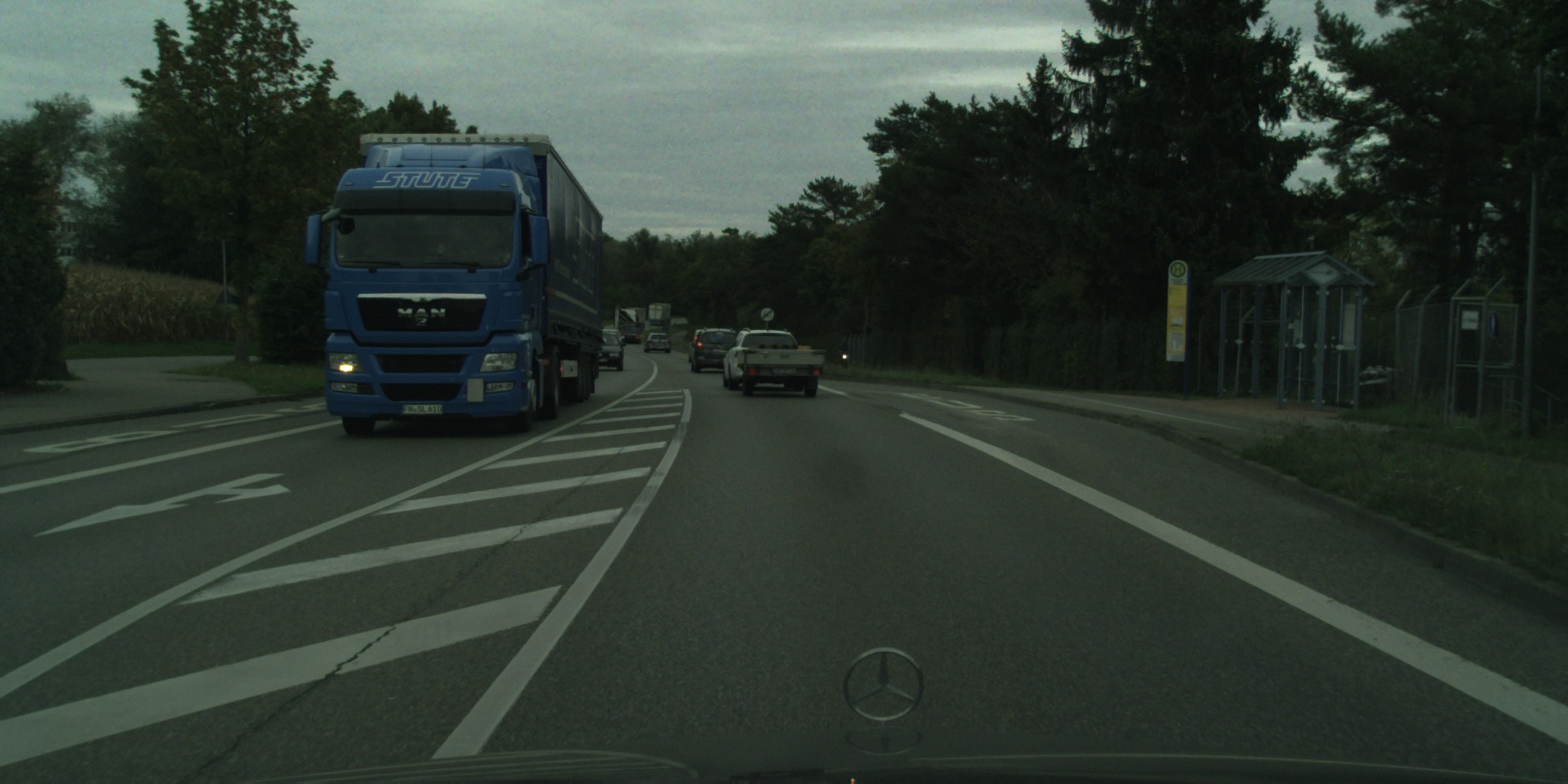} \\
  
\end{tabular}}
\caption{Qualitative results on GTA$\rightarrow$Cityscapes image translation. From top to bottom the rows show the source GTA image, CycleGAN translation, VSAIT translation, our DA-KPN translation, and a random Cityscapes real image as reference. DA-KPN's translations have significantly more noise than VSAIT, which more closely emulates the artifacts and saturation changes caused by glare as well as Gaussian noise in low lighting conditions.}
\label{fig:qualitative_translations_gta}
\end{figure*}

 We show qualitative results in Figure \ref{fig:qualitative_translations_gta} and zoomed-in results in Figure \ref{fig:crops}. Real images exhibit random noise in low lighting conditions (bottom right in Figure \ref{fig:qualitative_translations_gta}) and even color changes in shadow (Figure \ref{fig:crops}). While VSAIT is able to model decreased saturation well, it shows a loss of detail. DA-KPN translations exhibit more noise since noise is modeled explicitly. DA-KPN translations are generated at high resolution at much lower computational cost than VSAIT, leading to more fine-grained semantic details preserved from the source image, for example the scuffs in the pavement in Figure \ref{fig:qualitative_translations_gta}.
\par
\begin{figure*}[t]
\tabcolsep 0.03cm
\noindent\makebox[\textwidth]{
\begin{tabular}{ccccc}
  \includegraphics[trim = 0mm 0mm 0mm 0mm, clip, width=3cm, height=1.8cm]{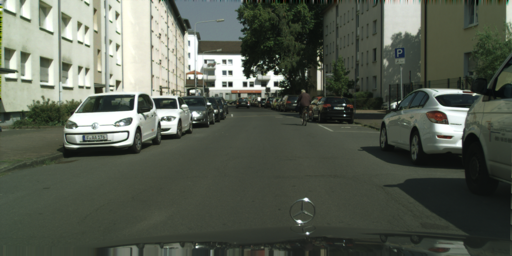} &
  \includegraphics[trim = 0mm 0mm 0mm 0mm, clip, width=3cm, height=1.8cm]{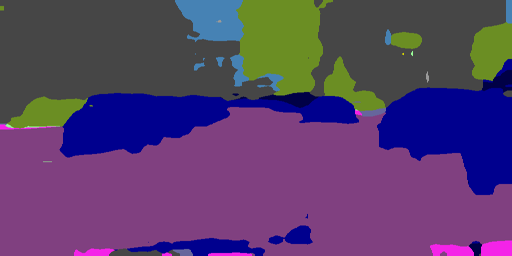} &
  \includegraphics[trim = 0mm 0mm 0mm 0mm, clip, width=3cm, height=1.8cm]{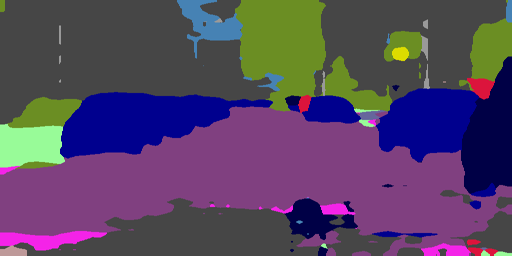} &
  \includegraphics[trim = 0mm 0mm 0mm 0mm, clip, width=3cm, height=1.8cm]{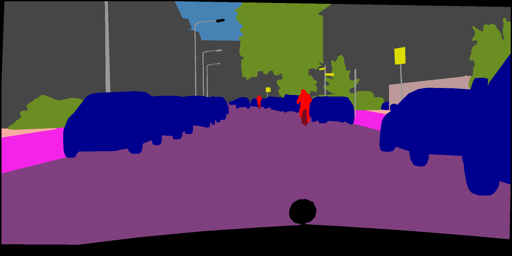} \\
  
  \includegraphics[trim = 0mm 0mm 0mm 0mm, clip, width=3cm, height=1.8cm]{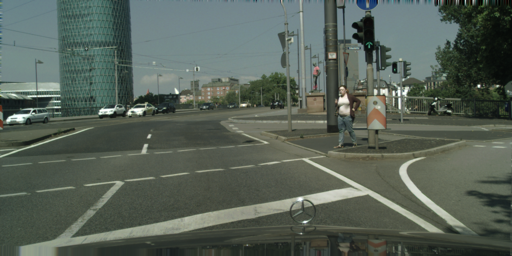} &
  \includegraphics[trim = 0mm 0mm 0mm 0mm, clip, width=3cm, height=1.8cm]{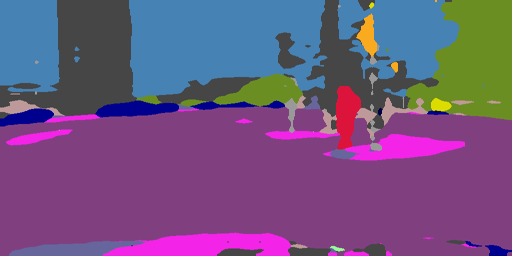} &
  \includegraphics[trim = 0mm 0mm 0mm 0mm, clip, width=3cm, height=1.8cm]{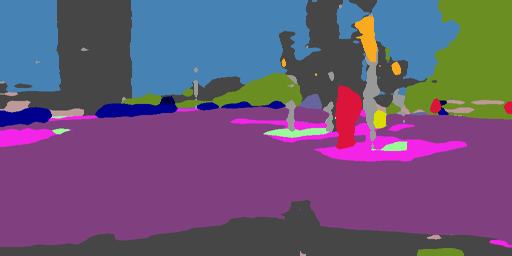} &
  \includegraphics[trim = 0mm 0mm 0mm 0mm, clip, width=3cm, height=1.8cm]{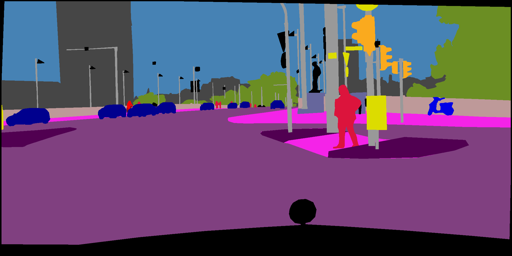} \\
  
  \includegraphics[trim = 0mm 0mm 0mm 0mm, clip, width=3cm, height=1.8cm]{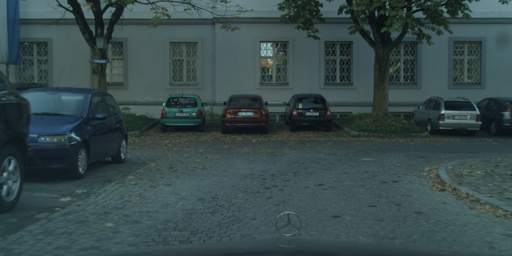} &
  \includegraphics[trim = 0mm 0mm 0mm 0mm, clip, width=3cm, height=1.8cm]{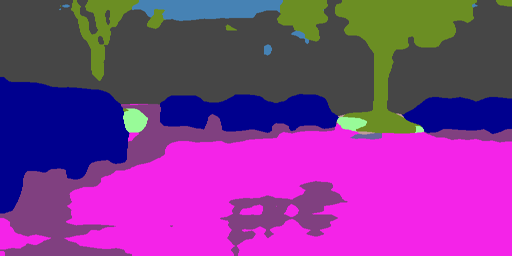} &
  \includegraphics[trim = 0mm 0mm 0mm 0mm, clip, width=3cm, height=1.8cm]{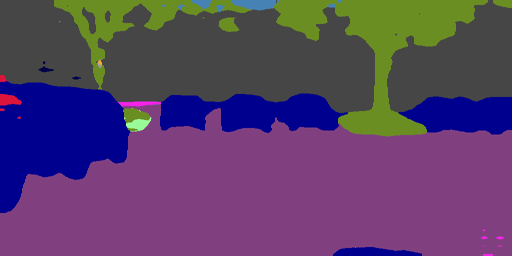} &
  \includegraphics[trim = 0mm 0mm 0mm 0mm, clip, width=3cm, height=1.8cm]{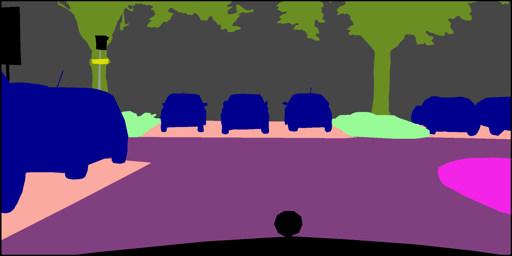} \\
  
  \includegraphics[trim = 0mm 0mm 0mm 0mm, clip, width=3cm, height=1.8cm]{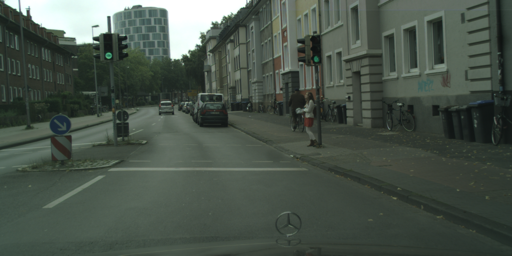} &
  \includegraphics[trim = 0mm 0mm 0mm 0mm, clip, width=3cm, height=1.8cm]{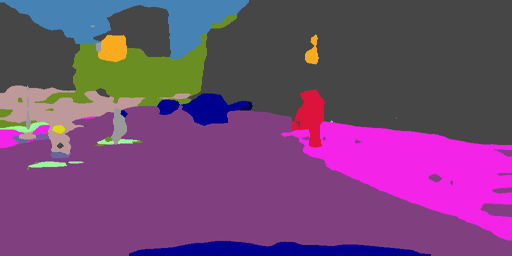} &
  \includegraphics[trim = 0mm 0mm 0mm 0mm, clip, width=3cm, height=1.8cm]{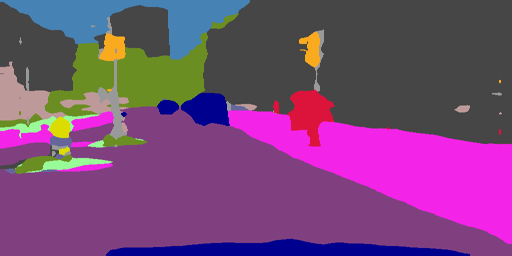} &
  \includegraphics[trim = 0mm 0mm 0mm 0mm, clip, width=3cm, height=1.8cm]{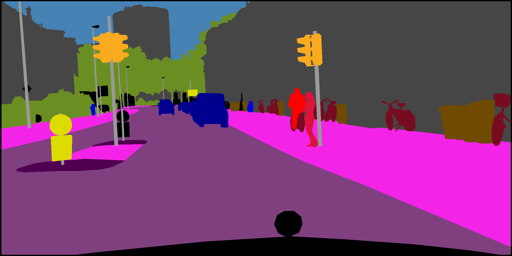} \\
  
  Target & VSAIT & \textbf{DA-KPN} & Ground Truth \\
\end{tabular}}
\caption{Segmentation predictions on Cityscapes from an FCN-Resnet50 trained on translations generated by DA-KPN and VSAIT.}
\label{fig:segmentations_cityscapes}
\end{figure*}

 The real image artifacts are most likely to affect the model's ability to distinguish boundaries of small objects far away, which are difficult to segment to begin with. This is confirmed when looking at the FCN-Resnet50's predictions on Cityscapes images in Figure \ref{fig:segmentations_cityscapes}: small objects such as street lights and distant people are under-segmented by the model trained on VSAIT translations in rows 1, 2, and 4. In row 3 the model must rely on structural information about roads and sidewalks to correctly identify the road since it is made of cobblestone instead of asphalt. Although there are no instances of this road texture in GTA, the noise added by DA-KPN helps the model become more robust to this texture change.

\begin{table}[t]
  \centering
  \caption{Face parsing results on CelebA by a an FCN-Resnet50 trained on FaceSynthetics translations.}
  \begin{tabular}{cccc}
    \toprule
    Translation Method & pxAcc & clsAcc & \textbf{mIOU} \\
    \midrule
    CycleGAN & 51.24 & 24.57 & 15.22 \\
    VSAIT & 77.12 & \textbf{41.41} & \textbf{29.80} \\
    DA-KPN-T (ours) & \textbf{81.37} & \underline{39.02} & \underline{29.11} \\
    \midrule
    {\color{gray} Target} & {\color{gray}95.90} & {\color{gray}85.66} & {\color{gray}79.17} \\
    \bottomrule
  \end{tabular}
  \label{tab:segmentation_results_celeba}
\end{table}

\begin{table}[t]
  \centering
  \caption{FCN-Scores reported on the FaceSynthetics translations.}
  \begin{tabular}{cccc}
    \toprule
    Translation Method & pxAcc & clsAcc & \textbf{mIOU} \\
    \midrule
    CycleGAN & 22.32 & 20.72 & 9.34 \\
    VSAIT & \textbf{25.21} & \textbf{18.91} & \textbf{10.44} \\
    DA-KPN-T (ours) & \underline{22.52} & \underline{18.48} & \underline{9.94} \\
    \bottomrule
  \end{tabular}
  \label{tab:fcn_scores_celeba}
\end{table}

\begin{table}
  \centering
  \caption{Transformation Component Ablation: Segmentation mIOU evaluated on Cityscapes validation set for FCN trained on DA-KPN's GTA translations.}
\begin{tabular}{ccccc}
    \hline
     & Full transformation &  -blur & -noise & -affine \\
    mIOU & \textbf{37.67} & 35.19 & 34.24 & 28.96 \\
    \hline
  \end{tabular}
  \label{tab:component_ablation}
\end{table}

\begin{figure*}[t]
\tabcolsep 0.03cm
\noindent\makebox[\textwidth]{
\begin{tabular}{ccccc}
  \includegraphics[trim = 0mm 0mm 0mm 0mm, clip, width=2.3cm]{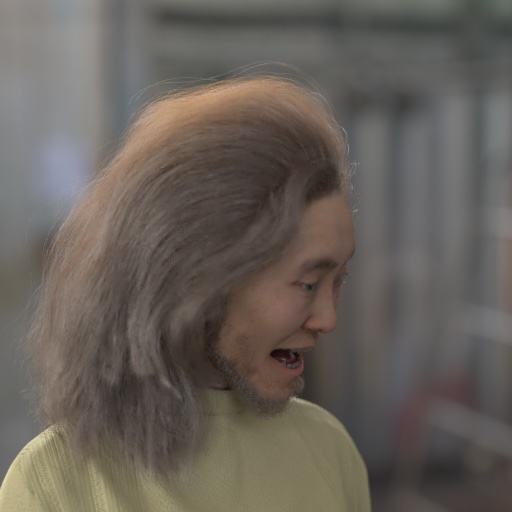} &
  \includegraphics[trim = 0mm 0mm 0mm 0mm, clip, width=2.3cm]{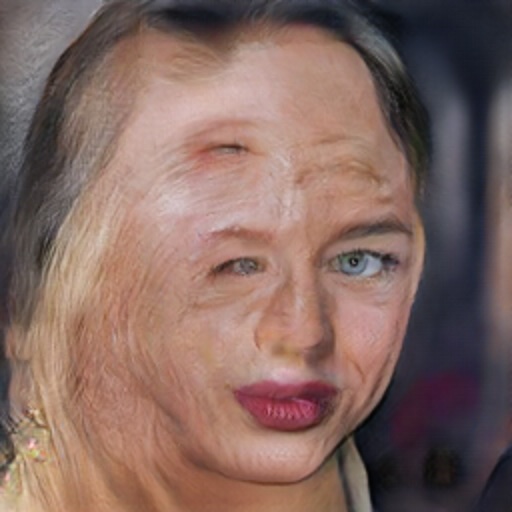} &
  \includegraphics[trim = 0mm 0mm 0mm 0mm, clip, width=2.3cm]{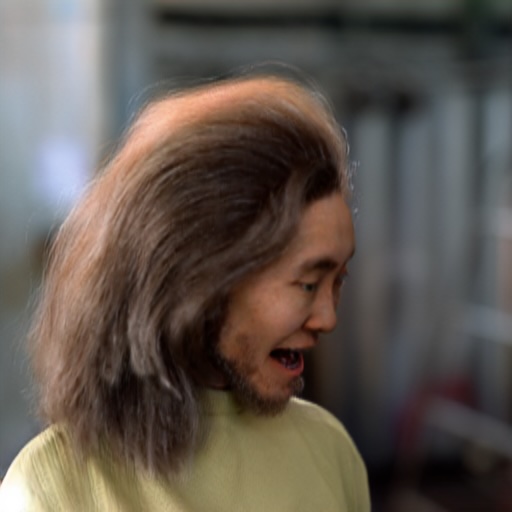} &
  \includegraphics[trim = 0mm 0mm 0mm 0mm, clip, width=2.3cm]{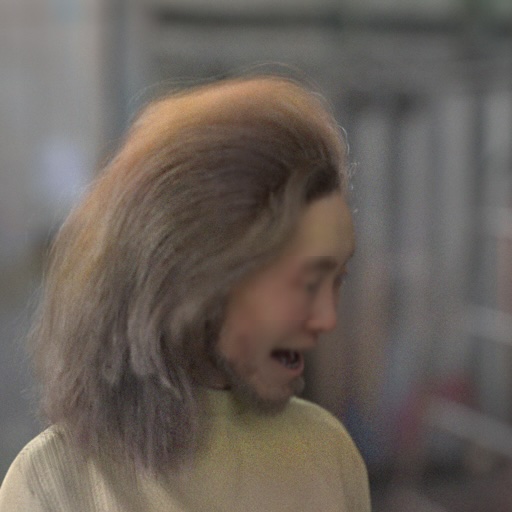} &
  \includegraphics[trim = 0mm 0mm 0mm 0mm, clip, width=2.3cm]{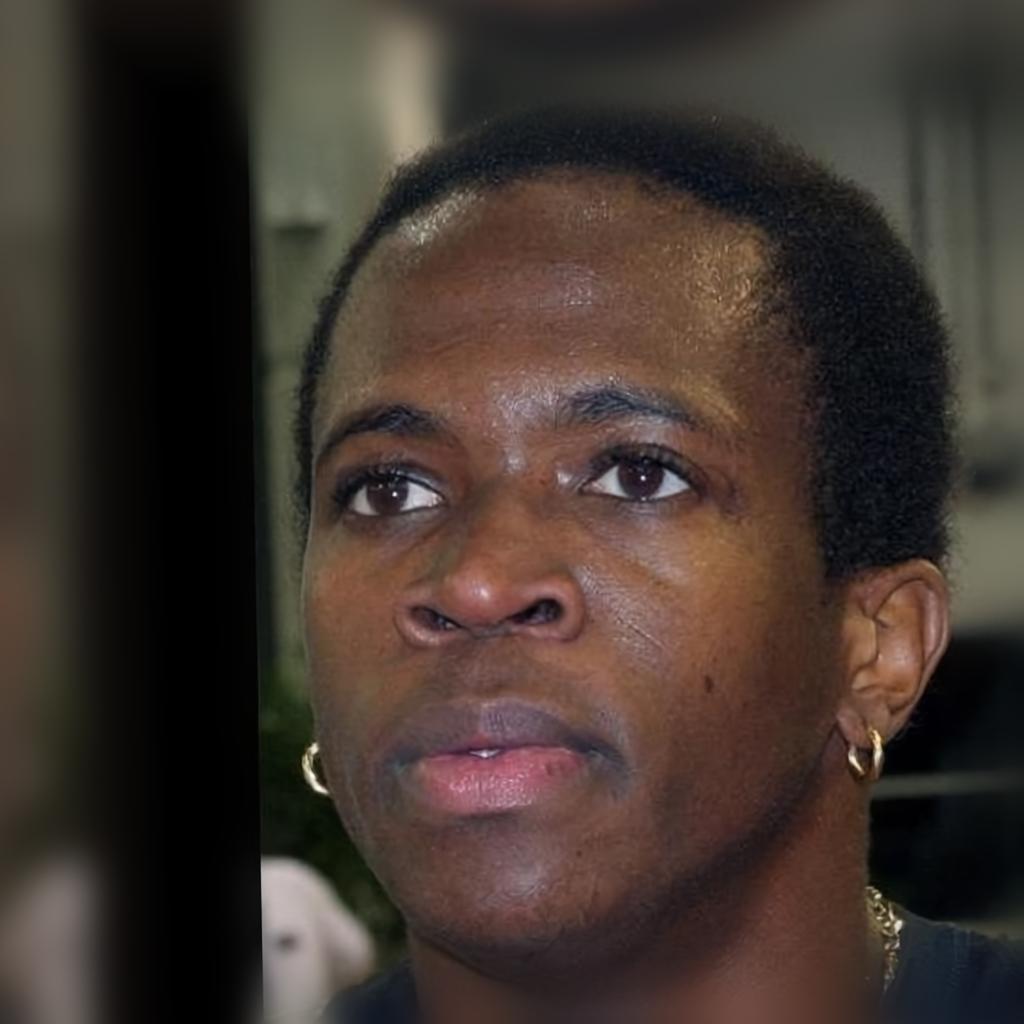} \\
  
  \includegraphics[trim = 0mm 0mm 0mm 0mm, clip, width=2.3cm]{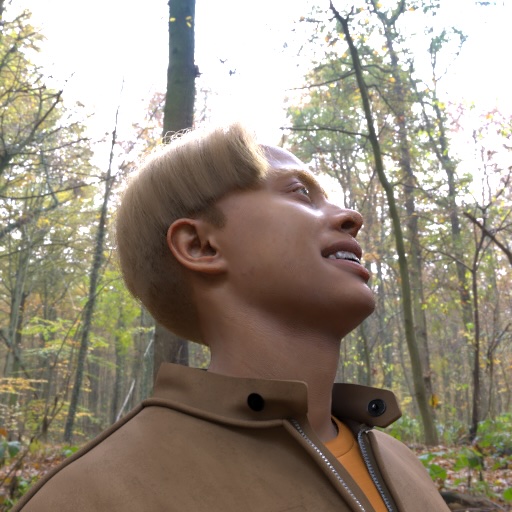} &
  \includegraphics[trim = 0mm 0mm 0mm 0mm, clip, width=2.3cm]{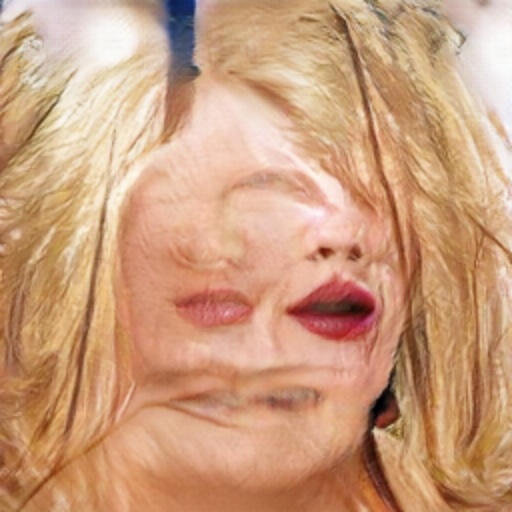} &
  \includegraphics[trim = 0mm 0mm 0mm 0mm, clip, width=2.3cm]{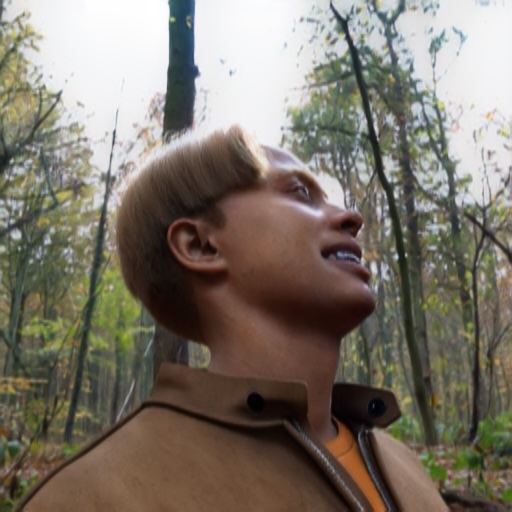} &
  \includegraphics[trim = 0mm 0mm 0mm 0mm, clip, width=2.3cm]{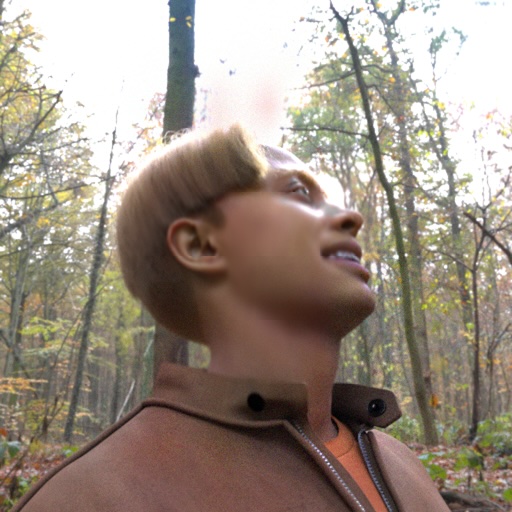} &
  \includegraphics[trim = 0mm 0mm 0mm 0mm, clip, width=2.3cm]{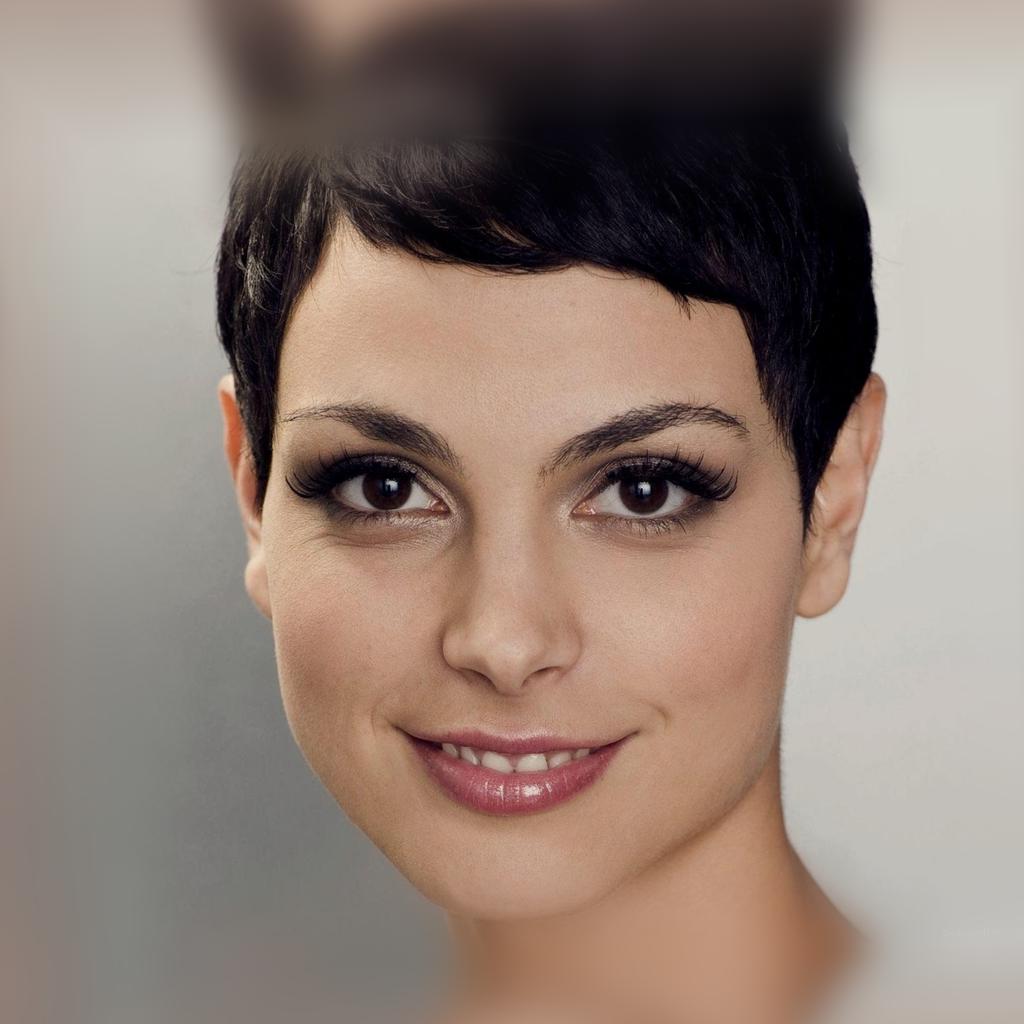} \\
  
  \includegraphics[trim = 0mm 0mm 0mm 0mm, clip, width=2.3cm]{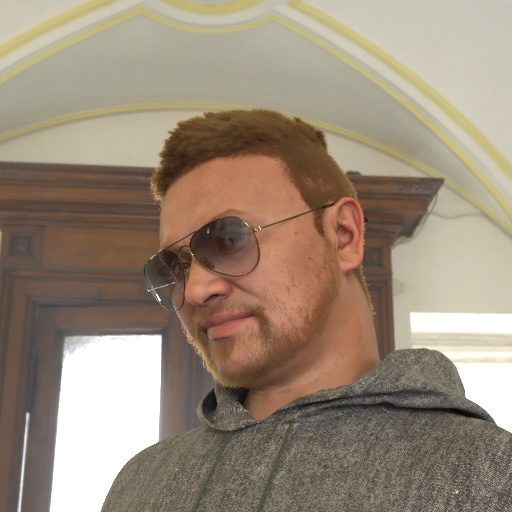} &
  \includegraphics[trim = 0mm 0mm 0mm 0mm, clip, width=2.3cm]{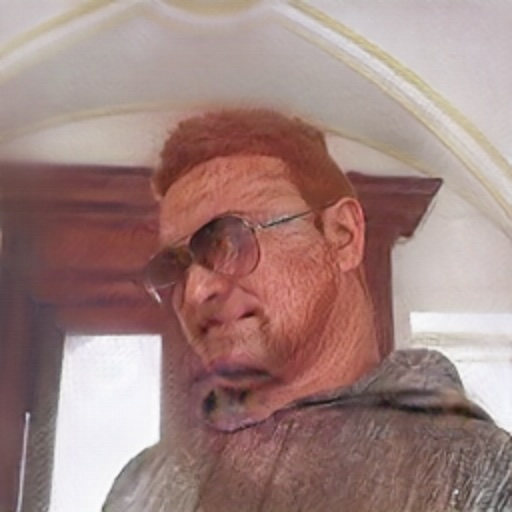} &
  \includegraphics[trim = 0mm 0mm 0mm 0mm, clip, width=2.3cm]{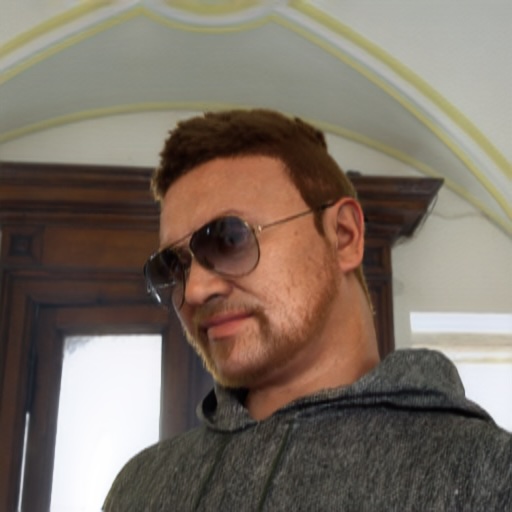} &
  \includegraphics[trim = 0mm 0mm 0mm 0mm, clip, width=2.3cm]{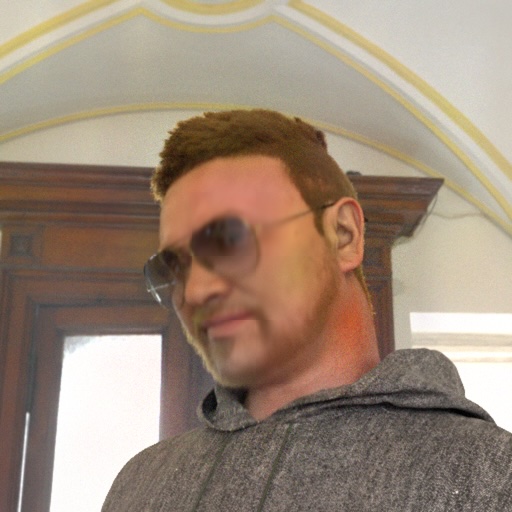} &
  \includegraphics[trim = 0mm 0mm 0mm 0mm, clip, width=2.3cm]{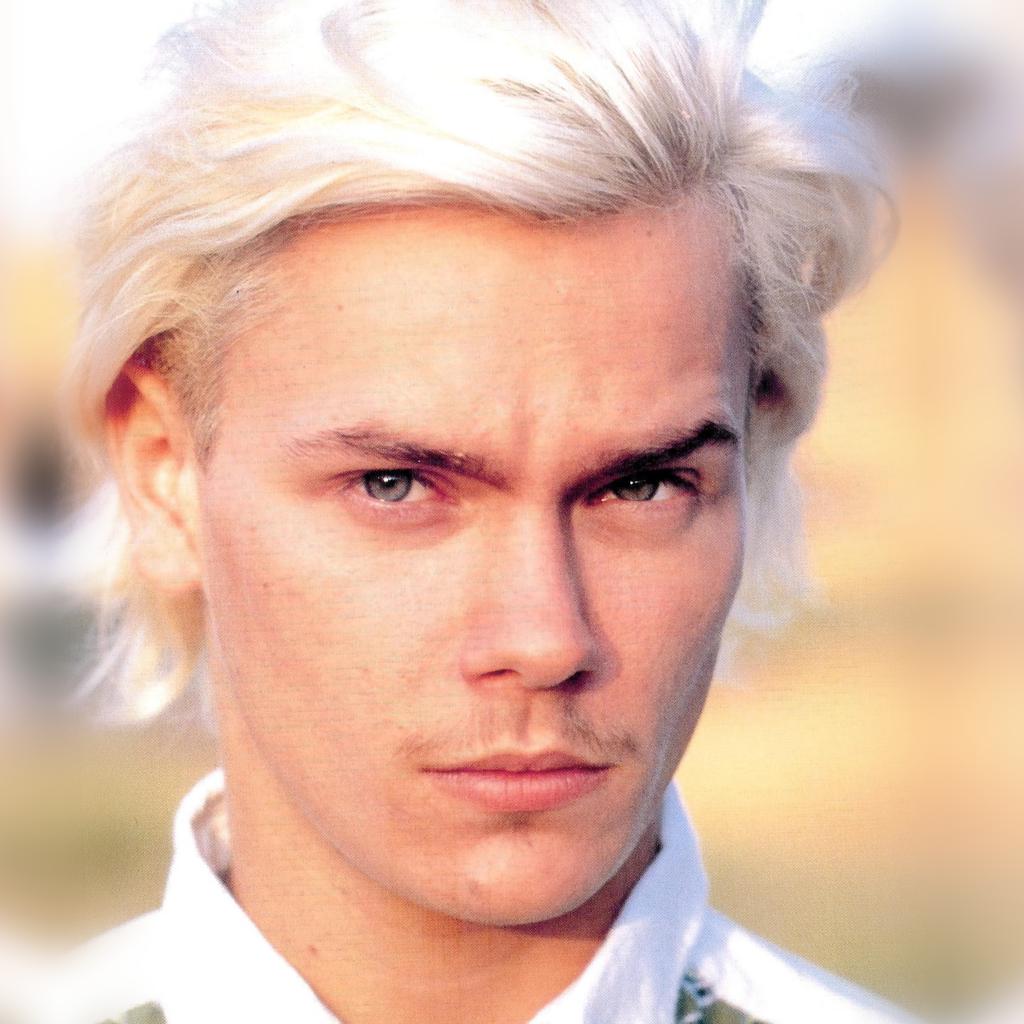} \\

  Source & CycleGAN & VSAIT & DA-KPN (ours) & Target \\
  
\end{tabular}}
\caption{Qualitative results on FaceSynthetics$\rightarrow$CelebA image translation. DA-KPN translations contain more Gaussian noise and blur than VSAIT due to our explicit modeling of these artifacts.}
\label{fig:qualitative_translations_face}
\end{figure*}
\begin{figure*}[t]
\tabcolsep 0.03cm
\noindent\makebox[\textwidth]{
\begin{tabular}{ccccccc}
  \includegraphics[trim = 0mm 0mm 0mm 0mm, clip, width=2cm, height=1.8cm]{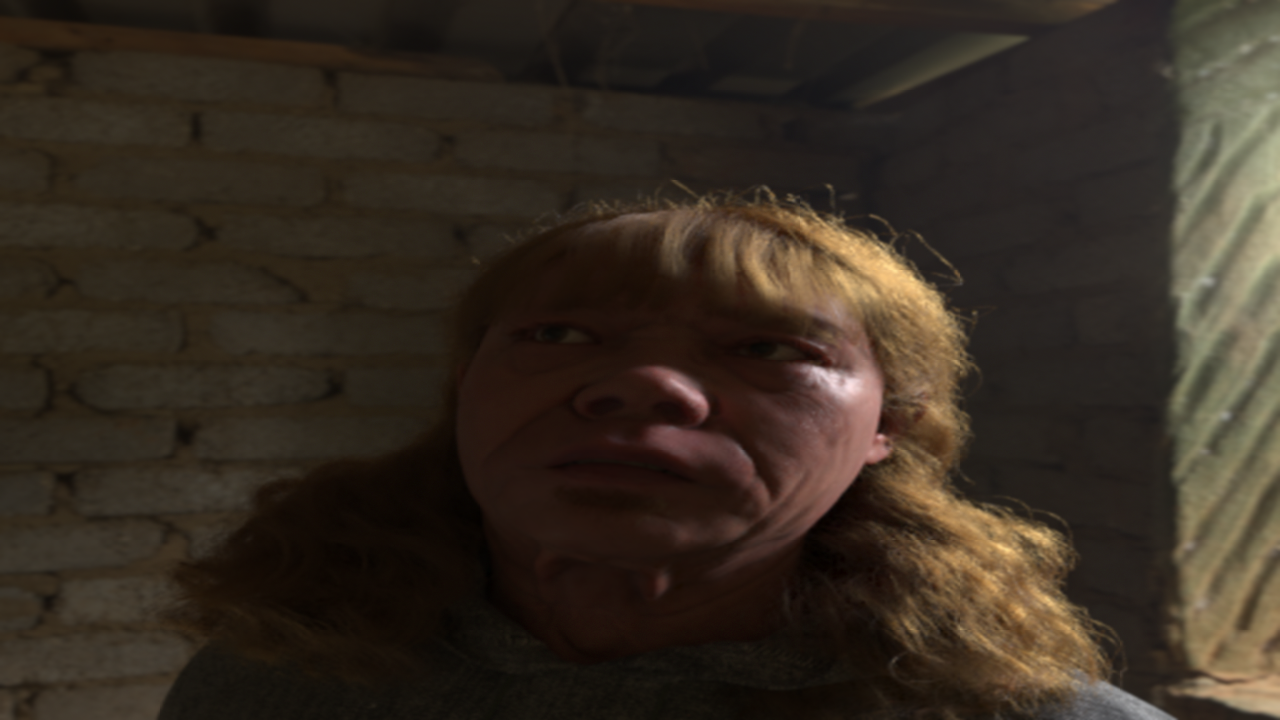} &
  \includegraphics[trim = 0mm 0mm 0mm 0mm, clip, width=2cm, height=1.8cm]{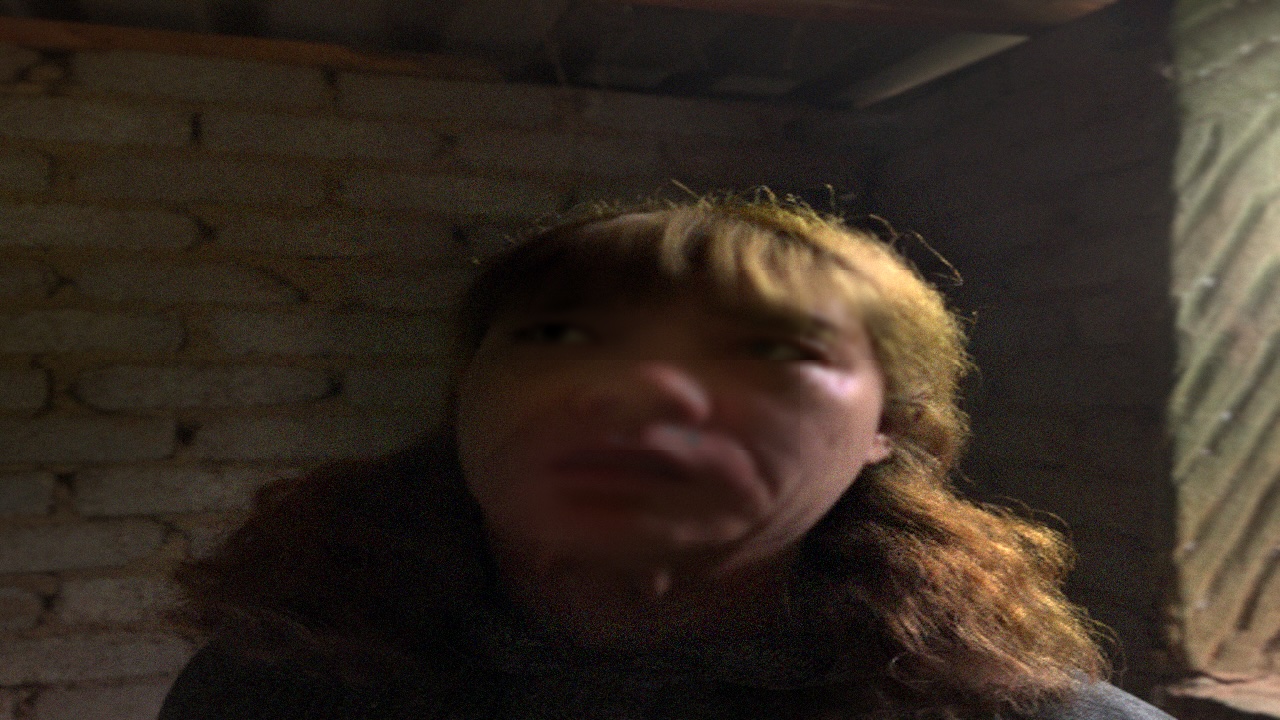} &
  \includegraphics[trim = 0mm 0mm 0mm 0mm, clip, width=2cm, height=1.8cm]{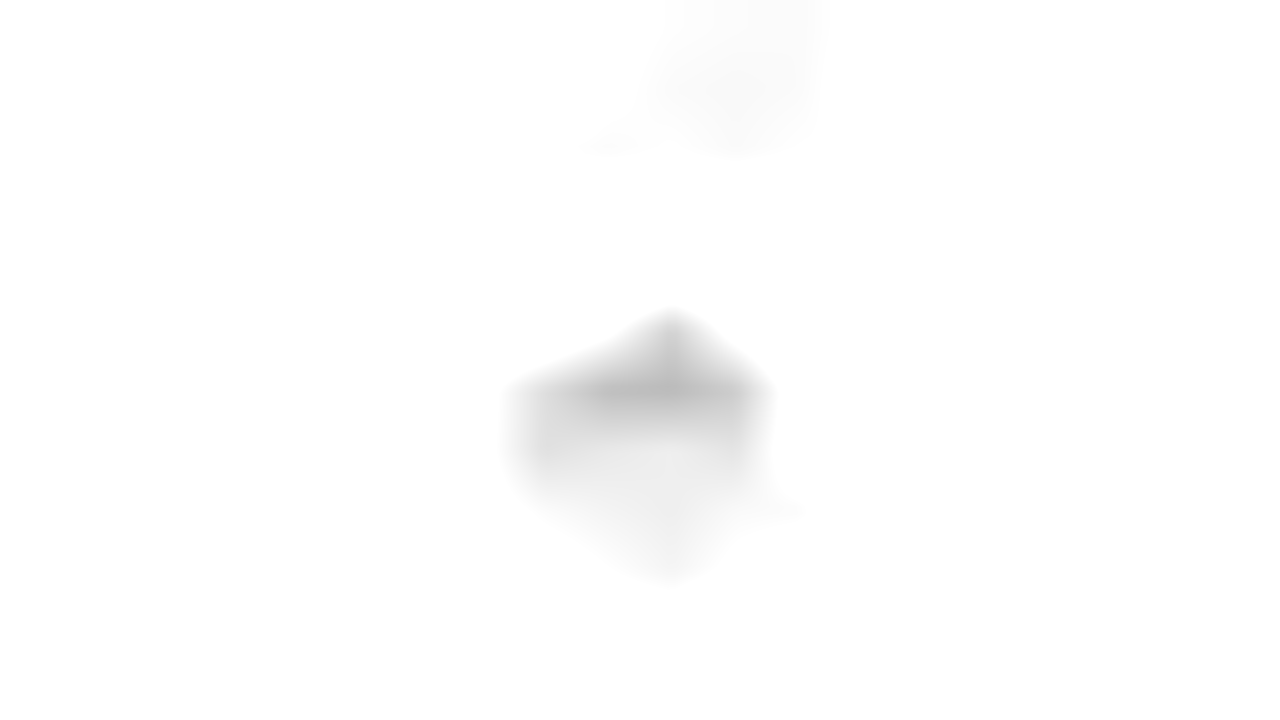} &
  \includegraphics[trim = 0mm 0mm 0mm 0mm, clip, width=2cm, height=1.8cm]{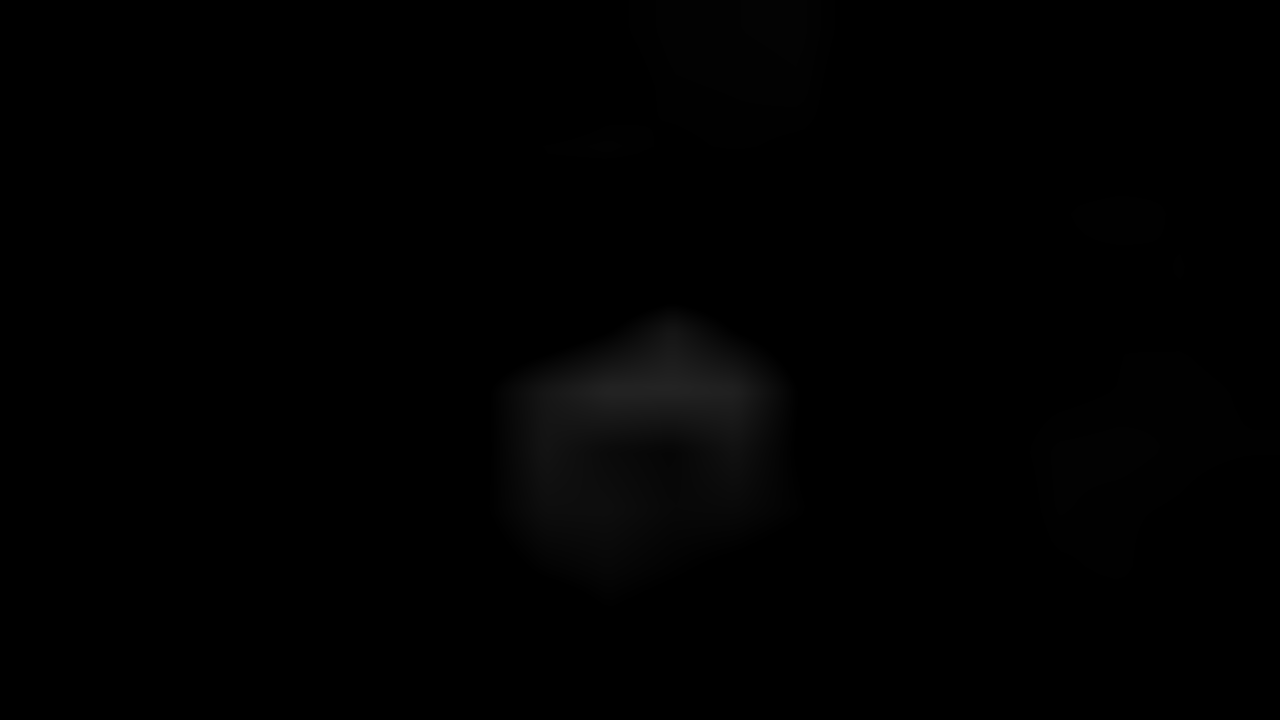} &
  \includegraphics[trim = 0mm 0mm 0mm 0mm, clip, width=2cm, height=1.8cm]{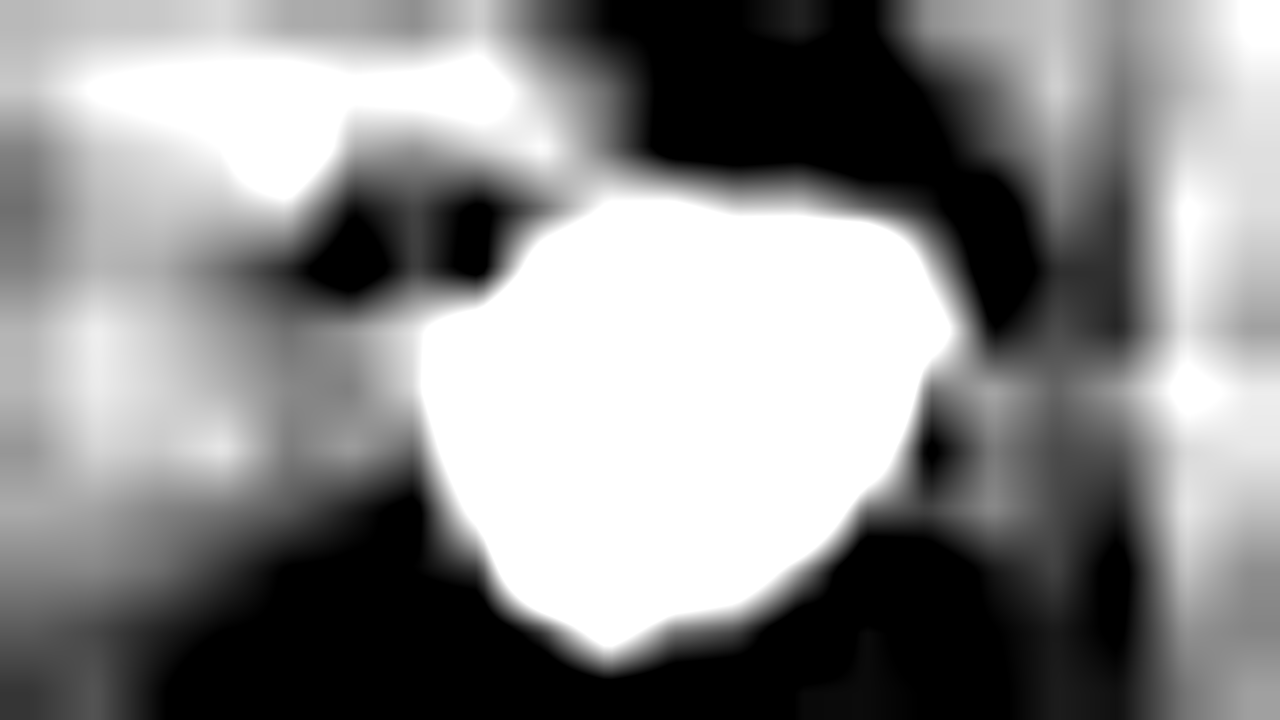} &
  \includegraphics[trim = 0mm 0mm 0mm 0mm, clip, width=2cm, height=1.8cm]{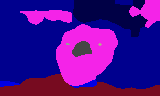} \\
  
  \includegraphics[trim = 0mm 0mm 0mm 0mm, clip, width=2cm, height=1.8cm]{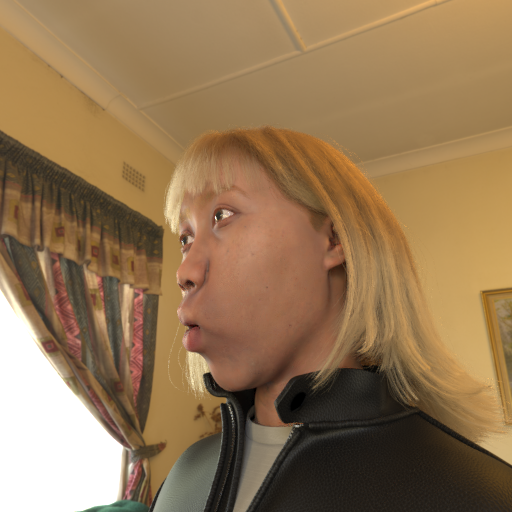} &
  \includegraphics[trim = 0mm 0mm 0mm 0mm, clip, width=2cm, height=1.8cm]{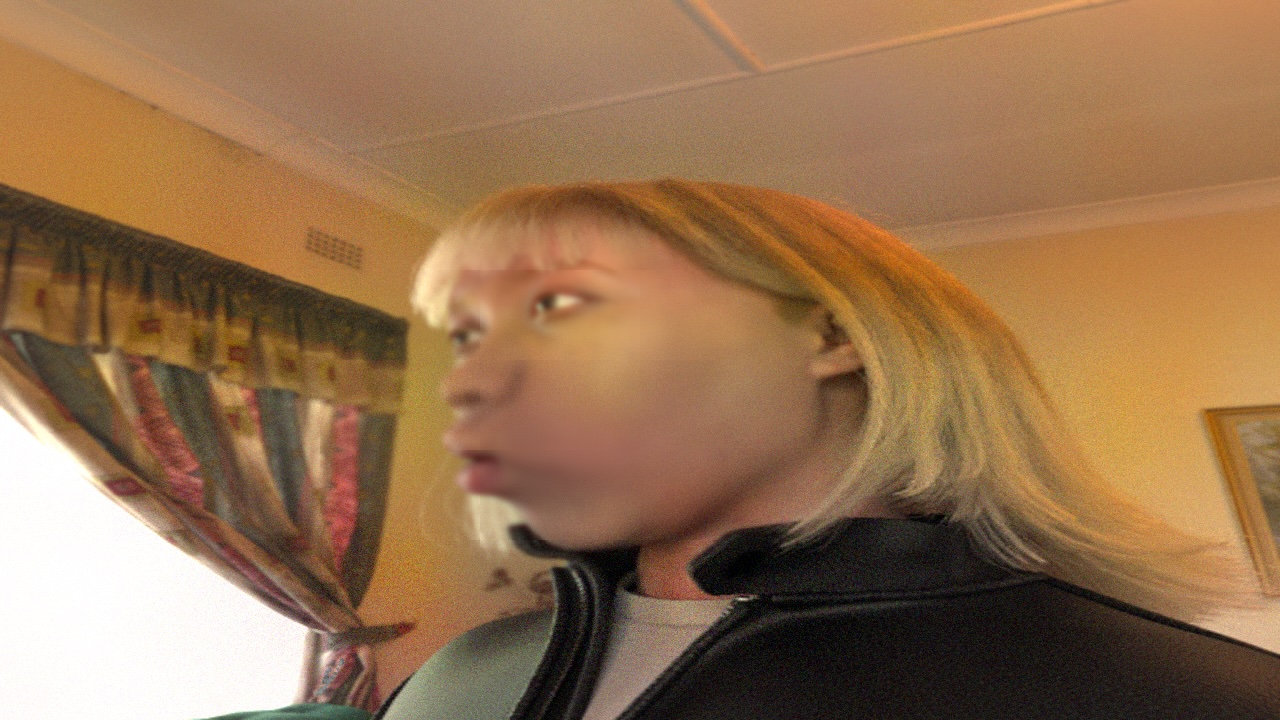} &
  \includegraphics[trim = 0mm 0mm 0mm 0mm, clip, width=2cm, height=1.8cm]{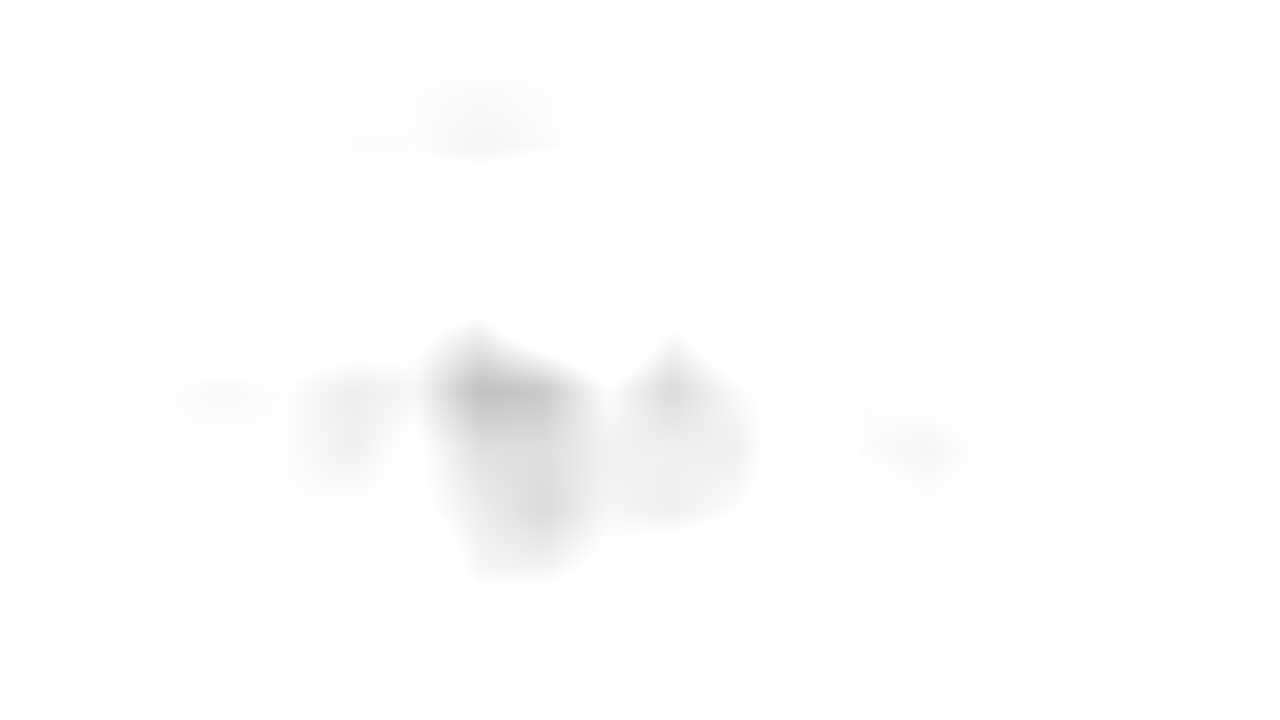} &
  \includegraphics[trim = 0mm 0mm 0mm 0mm, clip, width=2cm, height=1.8cm]{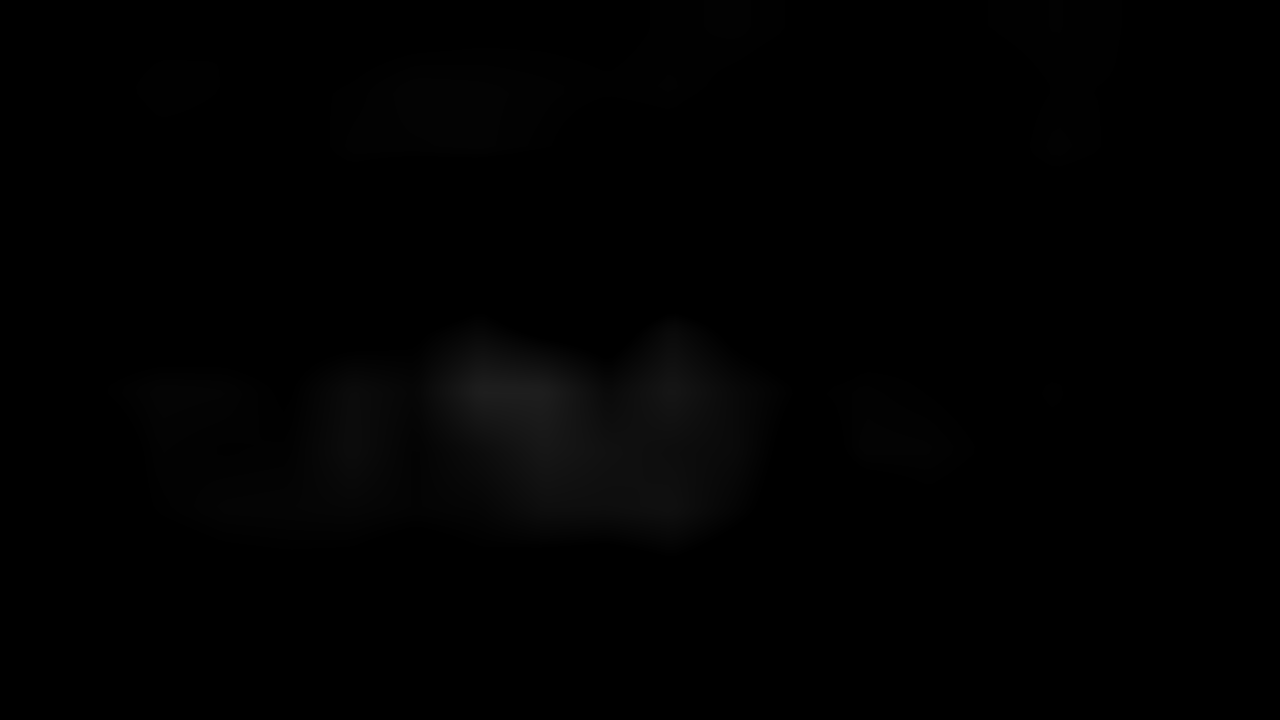} &
  \includegraphics[trim = 0mm 0mm 0mm 0mm, clip, width=2cm, height=1.8cm]{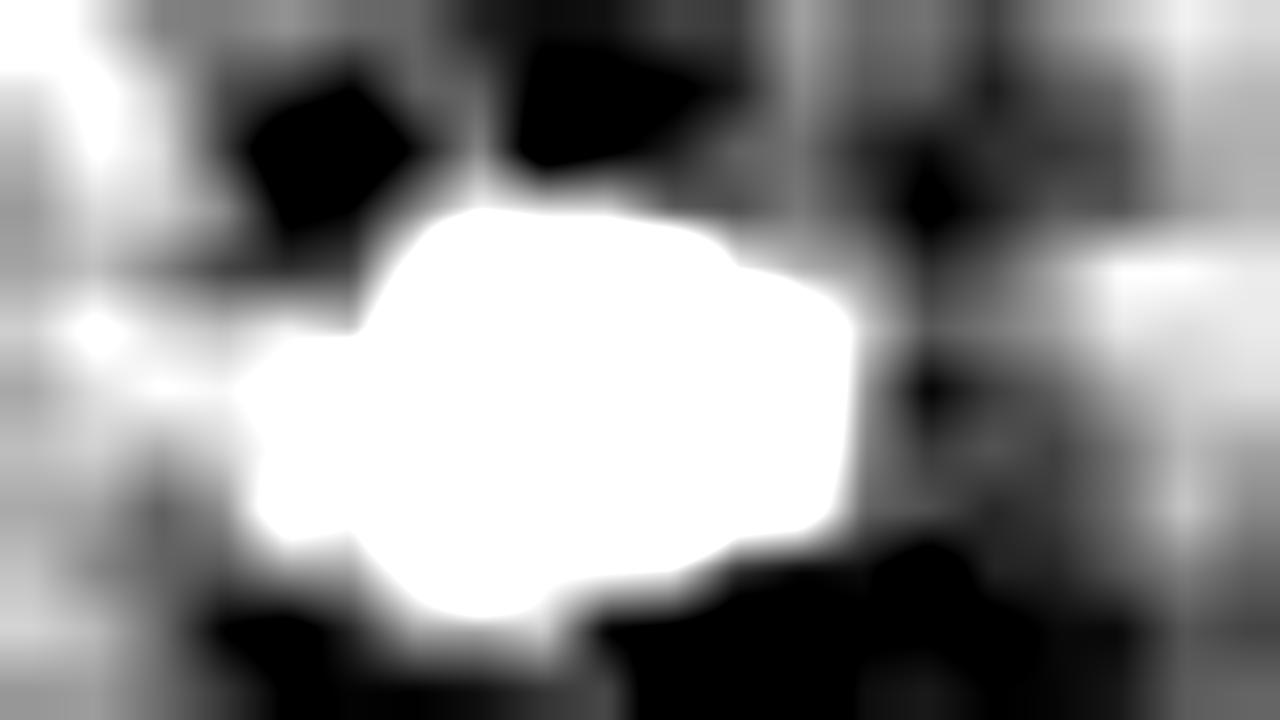} &
  \includegraphics[trim = 0mm 0mm 0mm 0mm, clip, width=2cm, height=1.8cm]{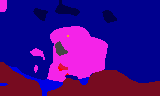} \\
\end{tabular}}
\caption{From left to right a Face Synthetics image, DA-KPN's translation, visualization of color weight and bias and blur (channels 2, 5 and 8), and the segmentation map generated by the semantic encoder.}
\label{fig:parameter_vis}
\end{figure*}
\subsection{Face Parsing}

\par Face parsing is a dense prediction task that is focused on classifying each pixel of a face as the part it belongs to. It can be implemented as semantic segmentation with the semantic classes corresponding to face parts instead of separate objects. For the source synthetic dataset we use FaceSynthetics \cite{Wood21}, which consists of 100,000 finely labeled images of synthetic people against real backgrounds. For the real target dataset we use CelebAMask-HQ \cite{Lee20a} (referred as CelebA) which consists of 30,000 finely labeled images across 19 semantic classes. There are 24,182 images in the training split and 2,993 images in the validation split that we use in our experiments. We process FaceSynthetics classes to align with those from CelebA. To the best of our knowledge we are the first to report face parsing results using unpaired image translation on these datasets.
\par We train DA-KPN on FaceSynthetics and CelebA using the same training process as for GTA$\rightarrow$Cityscapes. Then we generate translations on FaceSynthetics and use it to train an FCN-Resnet50 for face parsing on CelebA. We evaluate the network trained on these translations on the CelebA validation set in Table \ref{tab:segmentation_results_celeba}. We also train baselines CycleGAN and VSAIT on this pair of source and target datasets and report their results for comparison. DA-KPN outperforms CycleGAN and has comparable performance to VSAIT in mIOU and outperforms it on pixel accuracy. We compute FCN-Scores with retrained baselines in Table \ref{tab:fcn_scores_celeba} and see similar results.
\par Qualitative translation results are shown in Figure \ref{fig:qualitative_translations_face}. Ideally, the translation from synthetic to real faces should incorporate artifacts from real images, including lighting changes and motion blur. CycleGAN is not able to generate realistic translations due to the large domain shift between FaceSynthetics and CelebA, which includes changes in face angle, backgrounds and blur. VSAIT's translations are overly similar to the source except with higher saturation. Our DA-KPN's translations also follow the source input closely but Gaussian noise is visible and faces are blurred. Although parts of faces in CelebA may be blurred as shown in the right-most column, overblurring of face parts such as eyes and mouth likely contributes to a drop in our face parsing performance because it removes information about the boundaries of small parts.
\subsection{Ablation}
\par To show the contribution of each component of DA-KPN's transformation function, we evaluate the performance of an FCN trained on translations generated by DA-KPN when it is missing one of the affine, blur and noise components. In this experiment we use the DA-KPN-T model as our baseline, i.e. we pretrained the encoder on labeled training data from Cityscapes. Table \ref{tab:component_ablation} shows that removing any one of the affine, blur or noise components from the transformation function decreases performance.
\section{Discussion}
\par One advantage of DA-KPN is its interpretability. The semantic encoder can be used to understand how different regions undergo different transformations. We demonstrate this in Figure \ref{fig:parameter_vis} by showing, from left to right, a FaceSynthetics image, DA-KPN's translation, the KPN's parameter prediction at channels 2, 5 and 8 and lastly the semantic encoder's feature map. Channels 2 and 5 correspond to the weight and bias of the linear transformation in value space. Channel 8 corresponds to the blue channel's blur sigma parameter. The blur application mainly follows regions that are predicted as the skin class, most likely because heads exhibit many small movements in real-world scenarios. The color value is weighted higher in the skin region and the skin is brighter in the translation, matching CelebA's faces that tend to be illuminated with very bright lighting. The translations are consistent across samples: blur is applied to skin in both rows in Figure \ref{fig:parameter_vis}.
\par While DA-KPN's reliance on the encoder features brings interpretability to the translation, this makes the translation sensitive to errors in the feature map. In Figure \ref{fig:parameter_vis} there is an over-segmentation of the face in the sample on the bottom row, resulting in a brighter portion of the curtain in the background. This issue could be mitigated by applying the semantic encoder on higher resolution images which ensure that its features do not lose information during parameter prediction, at a slight increase in computation.
\par In our work we experimented with using either a source-pretrained or a target-pretrained semantic feature encoder. While DA-KPN's architecture flexibly allows different architectures for the semantic encoder, the outputs of the encoder must maintain a spatial resolution that allows transfer of detail from the input image pixels to parameter space. If the ratio of low resolution to high resolution is too high, then the upsampling process risks the same parameters being used for translation of different semantic classes and thus a suboptimal translation.

\section{Conclusion}

\par In this work we present a novel unpaired image translation method, DA-KPN, that focuses on closing the syn2real domain gap for segmentation. The core idea of DA-KPN is to use semantic features to predict pixel-wise parameters for a transformation function designed to target real image artifacts. By applying pixel-wise transformations to the source image, DA-KPN avoids structural changes endemic to GAN-based methods. An adversarial loss encourages both translated patches and the fully transformed image to resemble real images. We find that DA-KPN produces translations better suited for semantic segmentation than the state of the art. To the best of our knowledge we present the first unpaired image translation results for face parsing with FaceSynthetics$\rightarrow$CelebA. We demonstrate how each component of the translation function contributes positively to DA-KPN's performance. Finally, we demonstrate how the translation can be interpreted by visualizing the parameter channels output by the KPN. \newline


\par \noindent \textbf{Acknowledgment}
This work was supported by Electronics and Telecommunications Research Institute (ETRI) grant funded by the Korean government. [24ZB1200, Research of Human-centered autonomous intelligence system original technology] This work was also supported by the Institute of Information \& communications Technology Planning \& Evaluation(IITP) grant funded by the Korea government(MSIT) (No. RS-2024-00336738, Development of Complex Task Planning Technologies for Autonomous Agents, 100\%)

%
%
\bibliographystyle{splncs04}
\bibliography{main}
\end{document}